\documentclass[11pt]{article}
\usepackage[letterpaper,margin=1in]{geometry}
\usepackage{graphicx}
\usepackage{xcolor}
\usepackage[colorlinks=true,allcolors=blue]{hyperref}

\title{Characterizing Privacy Leakage in Spiking Neural Networks: A Systematic Analysis of Temporal Dynamics, Stochasticity, Neuronal Dynamics and Learning Mechanisms}
\author{Ayana Moshruba and Maryam Parsa\\
\small Electrical and Computer Engineering, George Mason University\\
\small Fairfax, Virginia, USA\\
\small \texttt{amoshrub@gmu.edu}}
\date{}

\begin{document}

\maketitle

\begin{abstract}

Spiking neural networks (SNNs) encode information not only in whether neurons fire but also in when they fire, and this temporal dimension shapes what a trained network reveals about its training data. Biologically inspired firing patterns can reduce membership leakage in SNNs without any change to the network architecture, which positions temporal dynamics as a design variable for privacy. A change in firing behavior, however, alters stochasticity, spike timing, neuronal response, and the learning problem at once, so harnessing this effect requires identifying which of these properties produces it. This study traces that origin by separating temporal representation, stochasticity, intrinsic neuron dynamics, and learning within a common experimental framework, evaluated under a black-box membership inference attack across image and tabular benchmarks. Varying the temporal representation shows that burst-like firing patterns reduce leakage relative to regular spiking, but at a cost in accuracy. Injecting randomness into the input, the spike timing, and the spike trains does not reproduce this reduction in leakage, which shows that the reduction arises from how spikes are arranged in time rather than from noise. The privacy benefit further depends on where these firing patterns originate. When neurons generate these patterns through their own adaptive dynamics, leakage falls while accuracy is preserved, so privacy no longer comes at the expense of performance. The learning algorithm then sets the baseline level of leakage. Under evolutionary training, the network already exhibits MIA performance close to 0.5, leaving temporal dynamics little room to improve privacy further. Privacy in SNNs is therefore not an inherent property of spikes but a product of how temporal activity is generated, processed, and learned, which offers a principled basis for designing neuromorphic systems that protect sensitive data.
\end{abstract}


\noindent\textbf{Keywords:} spiking neural networks, membership inference attacks, temporal dynamics, neuromorphic privacy, stochasticity, adaptive exponential integrate-and-fire neurons
\section{Introduction}

The growing use of machine learning in resource-constrained and privacy-sensitive environments has created a need for models that can balance computational efficiency with reliable protection of training data. Neuromorphic computing has received increasing attention in this context because it provides an event-driven alternative to conventional neural computation~\cite{davies2018loihi}. Spiking neural networks (SNNs), which form a central computational model within neuromorphic systems, process information through discrete spikes distributed over time. Their temporal mode of computation supports energy-efficient processing and allows information to be represented through both spike occurrence and spike timing. These properties have motivated the use of SNNs in edge intelligence \cite{yang2022lead}, autonomous systems \cite{zhu2024autonomous,halaly2023autonomous}, and healthcare \cite{tian2023neurocare,chu2022neuromorphic}, where computational efficiency and the local processing of potentially sensitive data can be important.

The privacy characteristics of SNNs have received increasing attention as these models have been considered for applications in which training data may contain sensitive or confidential information. Membership inference attacks (MIAs), which determine whether a particular sample was used during training, have become one of the principal tools for studying such leakage. Initial comparative studies suggested that SNNs can provide stronger resistance to membership inference than conventional Artificial Neural Networks (ANNs) across a range of datasets and learning settings \cite{moshruba2024neuromorphic}. This observation suggested that the event-driven and temporal nature of spike-based computation may influence how information about individual training samples is retained and exposed by the model.

Subsequent work, however, has shown that this privacy behavior cannot be characterized as a uniform property of SNNs. Membership leakage varies across static and neuromorphic datasets, and in some settings SNNs can be as vulnerable as, or even more vulnerable than, their ANN counterparts \cite{li2025membership}. The apparent privacy advantage also changes with simulation latency, which shows that the temporal extent of spike-based computation can directly affect membership inference vulnerability \cite{guan2025privacy}. At the same time, characteristics internal to the training process have been shown to alter the privacy-performance tradeoff. Surrogate-gradient selection and numerical quantization can significantly change membership leakage even when the broader network architecture remains unchanged \cite{moshruba2024neuromorphic}. These findings suggest that privacy in SNNs is not determined by the use of spikes alone, but by the particular computational and learning mechanisms through which spike-based representations are formed and optimized.

The role of temporal behavior becomes especially important in this context. Recent results have shown that biologically inspired firing patterns such as Poisson Burst and Delayed Burst can produce substantially different membership-inference behavior under the same underlying LIF architecture \cite{moshruba2025izhikevich}. These findings suggest that SNN privacy is not a fixed architectural property, but instead depends on how temporal information is represented, processed, and learned. What remains unclear is why these changes in temporal and computational behavior lead to different levels of membership leakage.

Temporal firing patterns provide a particularly useful setting for examining this question because they alter how the same input is represented over time while preserving the broader computational framework of the network. Regular spiking, bursting, and delayed responses differ in spike count, temporal organization, onset characteristics, and degree of stochasticity. These differences can change the information presented to the network without requiring a change in the underlying architecture. Previous observations showed that such changes in temporal representation are accompanied by measurable differences in the privacy-performance tradeoff \cite{moshruba2025izhikevich}. However, the observed relationship between firing behavior and privacy does not by itself identify the mechanism responsible for that variation.

The privacy characteristics associated with a given firing pattern may arise from several interacting sources, including stochasticity in spike generation, the temporal organization of the resulting spike trains, the intrinsic dynamics of the neuron model, and the learning process through which network parameters are optimized. Because these factors are closely coupled in SNN computation, changes in privacy cannot be attributed to temporal structure alone without further analysis. This motivates a controlled analysis that isolates stochasticity, neuron dynamics, and learning to determine how each shapes membership leakage and predictive performance. 

In this study, we first examine homogeneous and heterogeneous configurations of Rate, Poisson-Burst, and Delayed-Burst dynamics under a common LIF-based network to characterize how temporal organization affects the privacy-performance tradeoff. We then examine the role of stochasticity through temporal jitter, input noise, and spike-level noise, with the aim of determining how perturbations introduced at different stages of spike processing affect membership leakage. Next, we study intrinsic neuron dynamics using Adaptive Exponential Integrate-and-Fire neurons and compare privacy across firing behaviors that arise from the neuron model itself. Finally, we examine the influence of learning through gradient-based and evolutionary optimization, along with differentially private training. This progression provides a controlled basis for separating the effects of temporal representation, stochasticity, neuron dynamics, and learning, and for determining how their interaction shapes the privacy-performance behavior of SNNs.

The results show that the privacy behavior of SNNs cannot be attributed to temporal structure or stochasticity alone. Homogeneous and heterogeneous temporal dynamics alter the privacy-performance tradeoff, but increasing randomness through temporal jitter, input noise, or spike-level noise does not produce a consistent privacy gain and often degrades utility instead. Intrinsic neuron dynamics introduce a separate effect, as AdEx neurons reduce membership leakage while preserving or improving classification performance across the evaluated datasets. Learning further conditions the observed privacy behavior, since the same temporal dynamics exhibit different leakage characteristics under gradient-based, evolutionary, and differentially private training. These findings indicate that membership privacy in SNNs emerges from the interaction among temporal organization, stochasticity, neuron dynamics, and learning rather than from any one component in isolation. The contribution of this study is a systematic separation of these factors and an explanation of why biologically inspired temporal dynamics exhibit different privacy characteristics.
\section{Background \& Related Work}

\subsection{\textbf{Spiking Neural Networks}}

SNNs extend conventional neural computation by representing neuronal activity through discrete events distributed over time~\cite{tavanaei2019deep,auge2021survey}. Rather than propagating continuous-valued activations between layers, an SNN communicates through spike trains, where both the occurrence and timing of spikes can convey information~\cite{eshraghian2023training}. This temporal representation introduces an internal state that evolves throughout the simulation period, which distinguishes SNN computation from the predominantly static activation transformations used in conventional ANNs.

A spike train generated by a presynaptic neuron $j$ can be represented as a sequence of discrete firing events \cite{schmutz2020mesoscopic}. In continuous time, the spike train is commonly expressed as a sum of Dirac delta functions, \begin{equation} S_j(t)=\sum_k \delta(t-t_j^k), \end{equation}

where $t_j^k$ denotes the time of the $k$-th spike generated by neuron $j$, and $\delta(\cdot)$ denotes the Dirac delta function~\cite{schmutz2020mesoscopic}. A postsynaptic neuron receives weighted contributions from multiple presynaptic spike trains. A simplified representation of the synaptic input to neuron $i$ is

\begin{equation}
I_i(t)=\sum_j w_{ij}S_j(t),
\end{equation}

where $w_{ij}$ denotes the synaptic weight between presynaptic neuron $j$ and postsynaptic neuron $i$. The neuron integrates this input according to its internal state equations~\cite{eshraghian2023training}. The choice of neuron model therefore determines how incoming spikes modify the membrane state, when a spike is generated, and how the neuron behaves after firing.

\subsubsection{\textbf{Neuron Models in SNNs}}

The behavior of an SNN depends strongly on the neuron model used to represent membrane dynamics and spike generation. A neuron model determines how incoming synaptic activity is accumulated, how the internal neuronal state evolves over time, when an output spike is produced, and how the state changes after firing \cite{burkitt2006review,indiveri2011neuromorphic}. Different neuron models therefore provide different levels of computational complexity and biological realism, ranging from simple threshold-based formulations to adaptive dynamical systems capable of reproducing diverse firing behaviors \cite{izhikevich2003simple,indiveri2011neuromorphic}.

\paragraph{Leaky Integrate-and-Fire (LIF) Neuron.}
LIF neuron is one of the most widely used neuron models for representing spiking neuronal dynamics because it provides a simple and computationally efficient approximation of membrane integration and spike generation \cite{burkitt2006review,brette2005adaptive}. The membrane potential $V(t)$ evolves according to

\begin{equation}
\tau_m \frac{dV(t)}{dt}
=
-\left(V(t)-V_{\mathrm{rest}}\right)
+
R I(t),
\end{equation}

where $\tau_m$ is the membrane time constant, $V_{\mathrm{rest}}$ is the resting membrane potential, $R$ is the membrane resistance, and $I(t)$ denotes the input current \cite{burkitt2006review}. The neuron integrates incoming activity over time, while the leakage term causes the membrane potential to gradually return toward its resting state in the absence of sufficient stimulation \cite{burkitt2006review}.

A spike is generated when the membrane potential reaches the firing threshold $V_{\mathrm{th}}$,

\begin{equation}
S(t)=H\left(V(t)-V_{\mathrm{th}}\right),
\end{equation}

where $H(\cdot)$ denotes the Heaviside step function. Following a spike, the membrane potential is reset to $V_{\mathrm{reset}}$ \cite{burkitt2006review}. The LIF model captures the fundamental processes of integration, leakage, thresholding, and reset with relatively low computational complexity \cite{burkitt2006review,brette2005adaptive}. Its computational simplicity has made integrate-and-fire formulations particularly suitable for large-scale simulations and neuromorphic hardware implementations \cite{izhikevich2003simple,indiveri2011neuromorphic}. This simplicity, however, also limits the range of intrinsic firing behaviors that can be reproduced compared with more expressive neuron models \cite{brette2005adaptive,izhikevich2003simple}.

\paragraph{Izhikevich Neuron.}
The Izhikevich neuron model provides a biologically expressive representation of neuronal activity while retaining relatively low computational complexity \cite{izhikevich2003simple}. In addition to the membrane potential, the model maintains a recovery variable that represents slower ionic processes and provides negative feedback to the membrane potential \cite{izhikevich2003simple}. The interaction between these two state variables, together with a small set of model parameters, allows the neuron to reproduce a broad range of biologically observed firing behaviors, including regular spiking, fast spiking, intrinsic bursting, chattering, and low-threshold spiking \cite{izhikevich2003simple}. This flexibility makes the Izhikevich model useful for representing temporal firing behaviors that cannot be generated intrinsically by the standard LIF formulation.

The use of the Izhikevich model in gradient-based machine learning is less straightforward because its continuous membrane and recovery dynamics are interrupted by an event-triggered reset of both state variables after a spike \cite{izhikevich2003simple}. This reset introduces a discontinuity into the neuronal trajectory, while the spike-generation mechanism introduces the same non-differentiability that complicates gradient-based training in SNNs more generally \cite{neftci2019surrogate,bittar2022surrogate}. Consequently, direct application of conventional backpropagation through the exact spiking dynamics is nontrivial. Surrogate-gradient methods provide one practical solution by replacing the derivative of the non-differentiable spike function during the backward pass, which allows models with Izhikevich-type dynamics to be incorporated into gradient-based SNN training frameworks \cite{neftci2019surrogate,bittar2022surrogate}.

\paragraph{Adaptive Exponential Integrate-and-Fire Neuron(AdEx)}
AdEx neuron provides a biologically expressive alternative to simpler integrate-and-fire models. AdEx combines an exponential spike-initiation mechanism with an adaptation variable that represents the effect of previous neuronal activity on subsequent membrane dynamics \cite{brette2005adaptive}. The subthreshold membrane and adaptation dynamics are governed by a two-dimensional system of differential equations, while spike generation is followed by an explicit reset of the membrane potential and adaptation state \cite{brette2005adaptive,naud2008firing}. Appropriate parameter configurations allow the model to reproduce a diverse set of firing behaviors, including tonic spiking, adaptation, initial and regular bursting, delayed spiking, delayed bursting, transient spiking, and irregular spiking \cite{naud2008firing}. The model can also reproduce characteristics associated with regular-spiking pyramidal neurons and fast-spiking interneurons \cite{naud2008firing}.

The continuous subthreshold dynamics of AdEx are smooth, and the exponential term provides a gradual spike-initiation mechanism compared with the strict voltage threshold of the conventional LIF model \cite{brette2005adaptive}. The complete spiking model nevertheless remains non-differentiable because spike emission and state reset are discrete events \cite{bittar2022surrogate}. Gradient-based training therefore requires an approximation of the spike derivative, as it does for other deterministic spiking neuron models. Surrogate-gradient methods provide such an approximation and allow Backpropagation Through Time (BPTT) to be applied to SNNs with nonlinear and adaptive neuron dynamics \cite{neftci2019surrogate,bittar2022surrogate}. AdEx is therefore compatible with modern gradient-based SNN training when an appropriate surrogate treatment of spike generation is used, while retaining substantially richer intrinsic neuronal dynamics than the standard LIF formulation \cite{brette2005adaptive,naud2008firing,bittar2022surrogate}.

\subsubsection{\textbf{Learning in SNNs}}

Training SNNs requires the optimization of synaptic parameters while accounting for the temporal evolution of neuronal states and the discrete nature of spike generation \cite{neftci2019surrogate,eshraghian2023training}. The threshold operation that produces a spike is non-differentiable, which prevents the direct application of conventional gradient-based optimization through the exact spike function \cite{shrestha2018slayer,neftci2019surrogate}. Several learning strategies have therefore been developed to address the challenges associated with the binary and dynamical nature of SNNs \cite{neftci2019surrogate,eshraghian2023training}. In this work, the two learning paradigms of primary interest are gradient-based backpropagation and evolutionary optimization.

\paragraph{Backpropagation Through Time (BPTT) Learning.}
Gradient-based SNN training commonly relies on surrogate-gradient methods to overcome the non-differentiability of spike generation \cite{neftci2019surrogate,eshraghian2023training}. During the forward pass, spike generation remains discrete and follows the threshold dynamics of the neuron model. During the backward pass, the derivative of the non-differentiable spike function is replaced by a differentiable surrogate so that gradient information can propagate through the network \cite{neftci2019surrogate}. This approach preserves discrete spike generation during forward computation while providing an approximate gradient for parameter optimization.

For temporally simulated SNNs, surrogate-gradient learning can be combined with BPTT, where the recurrent neuronal dynamics are unfolded across the simulation horizon and gradients are propagated through the temporal evolution of the neuronal states \cite{neftci2019surrogate,eshraghian2023training}. Temporal backpropagation methods have also been developed specifically for SNNs. For example, SLAYER introduced a temporal credit-assignment mechanism that propagates errors through both layers and time while addressing the non-differentiability of the spike-generation function \cite{shrestha2018slayer}. These approaches permit SNNs to be optimized using gradient-based learning while retaining event-based spike generation during the forward simulation.

\paragraph{Evolutionary Learning.}
Evolutionary learning provides an alternative optimization strategy that does not require gradient information or differentiable neuron dynamics. Evolutionary Optimization for Neuromorphic Systems (EONS) applies evolutionary search directly to SNN design and training and can optimize network parameters and structure without differentiating through the spike-generation process \cite{schuman2020eons}. The optimization begins with a population of candidate SNNs, where each candidate is evaluated according to a task-specific fitness objective. Candidate networks are selected according to their fitness, and evolutionary operators such as mutation and crossover are applied to construct subsequent generations \cite{schuman2020eons,elbrecht2020evolving}. This process is repeated over successive generations until a predefined termination criterion is reached.

The absence of gradient requirements makes evolutionary optimization compatible with discrete spike events, recurrent network structures, and neuron models whose dynamics are difficult to optimize through conventional backpropagation \cite{schuman2020eons}. This flexibility also comes with a computational cost because multiple candidate networks must be evaluated repeatedly throughout the evolutionary process. The scalability of evolutionary SNN optimization can therefore become challenging as the task and network complexity increase \cite{elbrecht2020evolving}.

Backpropagation-based and evolutionary learning represent fundamentally different mechanisms for optimizing SNNs. Backpropagation relies on gradient information and surrogate approximations of spike generation, whereas evolutionary learning searches the parameter space through population-based optimization without requiring differentiability \cite{neftci2019surrogate,schuman2020eons}. These differences provide distinct learning conditions under which the predictive and privacy characteristics of SNNs can be examined.

\subsection{\textbf{Privacy in SNNs}}

Privacy research in SNNs has developed from two complementary questions: how privacy protection can be incorporated into spike-based learning, and whether the computational properties of SNNs themselves influence information leakage. The first question has motivated work on differential privacy, encrypted inference, and privacy-preserving distributed learning. The second has led to direct evaluations of SNNs under membership inference, model inversion, and gradient-based leakage attacks. The resulting literature shows that privacy in SNNs is shaped by several interacting design choices, including the temporal representation, neuron model, optimization procedure, numerical precision, simulation duration, and threat model.

Early studies primarily focused on introducing explicit privacy protection into SNN pipelines. PrivateSNN addressed data and class leakage during ANN-to-SNN conversion by replacing real conversion data with synthetic samples and subsequently adapting the converted model through temporal spike-based learning \cite{kim2022privatesnn}. The method reduced the forms of leakage considered in the study while preserving most of the predictive utility of the original model. DPSNN approached the problem through differential privacy and incorporated gradient perturbation directly into SNN training \cite{wang2022dpsnn}. Because privacy noise degraded predictive performance, the method introduced temporal enhanced pooling to recover part of the lost utility. A related approach, Encrypted-SNN, applied differential privacy during ANN-to-SNN conversion and showed that privacy protection could be incorporated into the conversion process without eliminating useful classification performance \cite{luo2023encryptedsnn}. This early body of work demonstrated that established privacy mechanisms could be adapted to spike-based learning, but it did not explain whether the internal computational properties of SNNs influence privacy independently of those mechanisms.

A separate line of research has examined privacy during inference through homomorphic encryption. Nikfam et al. evaluated encrypted SNN inference under the BFV scheme and showed that SNNs could retain competitive accuracy under encryption, although temporal simulation introduced additional computational cost \cite{nikfam2023homomorphic}. Li et al. subsequently developed a discretized SNN implementation under TFHE, where firing and reset operations were executed directly over encrypted data \cite{li2023tfhe}. The same direction was extended to convolutional SNNs through homomorphic implementations of LIF dynamics and parallelized bootstrapping \cite{li2023fhedicsnn}. Later frameworks such as SpyKing examined fully homomorphic SNN inference across multiple datasets and encryption configurations \cite{nikfam2025spyking}, while PrivSpike extended encrypted inference to deeper SNN architectures through CKKS-based computation and specialized approximations of LIF operations \cite{njungle2026privspike}. These approaches demonstrate that strong cryptographic protection is compatible with SNN inference, although their primary objective is to conceal the input during computation rather than to explain why a trained SNN leaks more or less information under a privacy attack.

Direct attack-based evaluations introduced a different perspective on SNN privacy. Model inversion studies showed that spike-based computation should not be assumed to prevent information recovery solely because spikes are discrete and non-differentiable. BrainLeaks introduced white-box inversion methods adapted specifically to SNNs and demonstrated that private information could still be reconstructed when the attack accounted for the structure of spike-based computation \cite{poursiami2024brainleaks}. Under black-box access, however, Poursiami et al. observed substantially stronger resistance for SNNs than for comparable ANNs \cite{poursiami2025spikes}. The difference between these findings illustrates an important characteristic of SNN privacy: the amount of information exposed by the model depends not only on the architecture, but also on the attacker's access and on how the attack interacts with temporal computation.

Membership inference has provided a broader framework for studying this dependence. Moshruba et al. compared SNNs and ANNs across multiple datasets, learning algorithms, software frameworks, and model configurations and reported lower membership leakage for SNNs across the evaluated settings \cite{moshruba2024neuromorphic}. The magnitude of this difference was not constant. Evolutionary learning produced different privacy behavior from gradient-based optimization, and differentially private training affected SNN and ANN utility differently under comparable privacy constraints. These results suggested that the observed privacy advantage was linked to the computational and learning configuration of the SNN rather than to the presence of spikes alone.

Later studies reinforced this view and showed that SNNs do not exhibit a uniform privacy advantage across all settings. Li et al. evaluated several MIAs across static and neuromorphic datasets and found that the relative vulnerability of SNNs depends strongly on the data modality and training configuration \cite{li2025membership}. SNNs could exhibit greater membership leakage than ANNs on neuromorphic datasets, while their relative behavior on static datasets varied across experimental conditions. The same study showed that ANN-to-SNN conversion and data augmentation could substantially change attack effectiveness. Guan et al. examined the temporal dimension more directly and found that membership-inference resilience decreased as simulation latency increased \cite{guan2025privacy}. An input-dropout strategy adapted to SNNs further improved attack effectiveness. These findings show that privacy can change even when the broad network family remains the same, which places greater importance on the internal temporal and learning configuration of the model.

This shift toward mechanism-level analysis has led to closer examination of the components that define SNN computation. Surrogate gradients are one such component because they determine how the non-differentiable spike function is approximated during gradient-based learning. Moshruba et al. showed that different surrogate-gradient formulations produce different levels of membership leakage and that numerical quantization further changes the privacy-performance tradeoff \cite{moshruba2025surrogate}. Privacy behavior has also been linked to the underlying neuron model. The Adaptive Differential Privacy SNN (ADPSNN) framework introduced an adaptive differential privacy mechanism based on the relationship between output spikes and class labels and reported different utility characteristics for LIF and IF neurons under privacy perturbation \cite{luo2025adpsnn}. The Relevance-based Adaptive Differential Private Spiking Neural Network (RADP-SNN) framework subsequently used neuron-level relevance to determine the allocation of differential privacy noise and similarly found that the resulting privacy-utility behavior depended on the underlying neuron model \cite{liu2026radpsnn}. Differentially private SNNs have also been studied in application specific settings such as social robotics, where privacy protection was evaluated alongside robustness and resource efficiency \cite{aitsam2025dpsnn}. Privacy analysis has further been extended to graph structured spiking models through membership inference and structural differential privacy \cite{moshruba2025edgeprivate}. These results show that privacy mechanisms interact with the structure and learning dynamics of the SNN rather than acting independently of them.

Temporal representation introduces another important source of variation. Biologically inspired firing patterns can change the spike sequence presented to the network without requiring any modification to the underlying architecture or neuron model. Moshruba et al. examined Poisson Burst and Delayed Burst dynamics within LIF-based SNNs and found that the resulting temporal firing patterns produced distinct privacy-performance tradeoffs under membership inference \cite{moshruba2025izhikevich}. Poisson Burst preserved utility close to conventional rate dynamics while reducing membership leakage, whereas Delayed Burst produced stronger privacy in several settings at a larger utility cost. This result is particularly relevant because the observed privacy differences arise under the same LIF architecture. The temporal dynamics itself therefore becomes a source of measurable privacy variation. At the same time, burst-based dynamics modify several properties simultaneously, including spike count, timing, onset, stochasticity, and temporal organization. The observed privacy effect cannot therefore be assigned to any one of these characteristics without further analysis.

The interaction between privacy and SNN computation has also been examined in federated settings. Han et al. combined federated SNN learning with differential privacy at both the client and server levels and showed that privacy perturbation could reduce membership and property inference while preserving the energy advantages of spike-based computation \cite{han2023federated}. More recent work has examined the information contained in the shared gradients themselves. Aksu et al. adapted gradient inversion attacks to federated SNNs and found substantially weaker reconstructions than those obtained from ANN gradients under the evaluated conditions \cite{aksu2025federated}. Their analysis associated this behavior with the event-driven forward process and surrogate-gradient training, which again links privacy to the mechanism through which the model represents and propagates information.

Privacy mechanisms can also alter the temporal signals on which SNN learning depends. Pereira et al. showed that differential privacy operations such as clipping and noise injection systematically change firing rate statistics in federated SNNs, with corresponding effects on aggregation signals and client-selection stability \cite{pereira2026sensitivity}. This result is important because the relationship operates in both directions: temporal behavior can influence privacy, and privacy protection can modify the temporal behavior of the network. Recent federated frameworks have started to account for this interaction explicitly. The Privacy-Preserving and Communication-Efficient Federated Spiking Neural Learning framework, PriSparK, combines event-sparse updates, differential privacy, error feedback, and heterogeneity-aware aggregation to address privacy and communication efficiency within a unified federated SNN framework \cite{liu2026prispark}. A distributed Integrated Sensing and Communication (ISAC) framework similarly combines federated SNN learning, secure aggregation, adaptive differential privacy, and spike-aware temporal processing for privacy-sensitive edge sensing \cite{alam2026federated}. These studies extend the privacy problem from individual SNN models to distributed learning systems in which temporal activity, privacy protection, communication, and optimization are closely connected.

The current literature therefore supports a more nuanced interpretation of SNN privacy than the early question of whether SNNs are inherently more private than ANNs. Explicit protection mechanisms such as differential privacy, homomorphic encryption, and secure aggregation can reduce information exposure, but their utility and effectiveness depend on the computational structure of the SNN. Direct attack evaluations further show that SNNs can exhibit privacy behavior that differs substantially from ANNs, but the direction and magnitude of that difference depend on the attack model, dataset, simulation latency, optimization procedure, and model configuration \cite{poursiami2024brainleaks,moshruba2024neuromorphic,li2025membership,guan2025privacy,poursiami2025spikes,aksu2025federated}. Studies of surrogate gradients, neuron models, temporal dynamics, and firing-rate encoding provide additional evidence that privacy is closely linked to the internal mechanisms of spike-based computation \cite{moshruba2025surrogate,moshruba2025izhikevich,luo2025adpsnn,liu2026radpsnn,pereira2026sensitivity}.

An important gap nevertheless remains. Most existing studies isolate one factor at a time, such as latency, surrogate gradients, quantization, neuron models, differential privacy, or temporal dynamics. As a result, the literature identifies many conditions under which SNN privacy changes, but it does not explain how these conditions relate to one another or which mechanisms are responsible for the observed differences. This limitation is particularly important for biologically inspired temporal dynamics, where changes in firing behavior can simultaneously alter stochasticity, spike timing, burst structure, neuronal response, and the learning process. A systematic analysis that separates these effects is therefore necessary to determine where the corresponding privacy behavior originates and how each component contributes to the resulting privacy-performance tradeoff.

\subsection{\textbf{Membership Inference Attacks(MIAs)}}

MIAs evaluate whether a trained machine learning model reveals information about the individual samples used during training. Given a target sample and access to a trained model, the adversary attempts to determine whether that sample belonged to the model's training set. The problem can be formulated as a binary inference task. For a target model $f_{\theta}$ trained on dataset $\mathcal{D}_{\mathrm{train}}$, an attack $\mathcal{A}$ predicts a membership variable

\begin{equation}
\hat{m}
=
\mathcal{A}\left(x,y,f_{\theta}(x)\right),
\qquad
\hat{m}\in\{0,1\},
\end{equation}

where $\hat{m}=1$ denotes a predicted member and $\hat{m}=0$ denotes a predicted non-member. Depending on the threat model, the adversary may observe the predicted class, confidence scores, loss values, intermediate representations, gradients, or model parameters. The objective remains the same: to distinguish samples that influenced model training from samples that did not \cite{shokri2017membership,yeom2018privacy}.

The privacy significance of membership inference depends strongly on the application. Knowledge that a particular record was used to train a model may itself reveal sensitive information when membership in the training set is associated with a medical condition, financial status, demographic group, or other confidential attribute. The attacker does not need to reconstruct the original record for a privacy violation to occur. The membership decision alone can disclose whether an individual contributed data to a sensitive dataset \cite{shokri2017membership}. MIAs have consequently become a widely used empirical framework for measuring whether information about individual training samples remains detectable through a trained model.~\cite{song2021systematic,carlini2022membership}.

The modern formulation of membership inference was introduced by Shokri et al. \cite{shokri2017membership} under a black-box threat model. Their attack exploits the observation that a model can behave differently on samples that participated in training and samples that were not observed during optimization. To learn this distinction, the adversary trains a collection of \textit{shadow models} on datasets drawn from a distribution similar to that of the target model. Because the attacker controls the training sets of the shadow models, the membership status of each sample is known. Prediction outputs from member and non-member samples are then used to train an attack classifier, which learns to distinguish the response patterns associated with the two groups. The resulting classifier is applied to the outputs of the target model to infer the membership status of previously unseen samples. This formulation established that model outputs alone can reveal information about the composition of the training set, even when the attacker has no direct access to the target model's parameters or training data.

Later work showed that successful membership inference does not necessarily require the complete shadow-model setting assumed in the original attack. Yeom et al. demonstrated that the loss of a target model on an individual sample can itself provide a membership signal and established a formal connection between membership leakage and generalization behavior \cite{yeom2018privacy}. Their analysis showed that overfitting is sufficient for membership inference vulnerability, although it is not necessary. Salem et al. subsequently relaxed several assumptions of the Shokri attack and showed that effective membership inference can be performed with fewer shadow models, limited knowledge of the target architecture, and auxiliary data that do not precisely match the target training distribution \cite{salem2019mlleaks}. These developments established that membership leakage is not restricted to adversaries with extensive knowledge of the target training process.

Research on MIAs also expanded the range of information available to the attacker. While the original Shokri attack relies on black-box prediction outputs, white-box attacks use internal information such as model parameters, intermediate activations, gradients, or parameter updates. Nasr et al. showed that such internal signals can expose additional membership information and developed white-box attacks for both centralized and federated learning settings \cite{nasr2019comprehensive}. At the opposite end of the access spectrum, Choquette-Choo et al. demonstrated that confidence scores are not strictly necessary for membership inference \cite{choquettechoo2021label}. Their label-only attacks infer membership from the stability of the predicted class under input perturbations. This result showed that restricting an application programming interface to return only class labels does not necessarily eliminate membership leakage.

The methodology used to evaluate MIAs has also evolved. Early attacks commonly reported aggregate attack accuracy or related average-case measures, but subsequent studies showed that such metrics can obscure substantial differences in privacy risk across individual samples. Song and Mittal developed a systematic evaluation framework that compared multiple attack signals and emphasized the need to assess privacy leakage beyond a single attack classifier or aggregate accuracy value \cite{song2021systematic}. Carlini et al. later formulated membership inference as a per-example hypothesis-testing problem and introduced the Likelihood Ratio Attack (LiRA) \cite{carlini2022membership}. LiRA estimates how likely an observed model response is under the hypotheses that a sample was included or excluded from training. Their analysis also emphasized attack performance at very low false-positive rates, where an attacker can identify a smaller number of members with substantially greater confidence. These advances demonstrate that membership inference is not a single fixed attack, but a family of inference procedures that differ in adversarial access, assumptions, attack signals, and evaluation objectives.

Despite these methodological differences, the central principle remains consistent: membership inference measures whether the behavior of a trained model contains a detectable statistical distinction between training samples and samples that were not used for training. This property makes MIA particularly useful as an empirical privacy measure when the objective is to compare how changes in model design or training affect information leakage. In this study, we use the original black-box shadow-model attack introduced by Shokri et al. \cite{shokri2017membership}. A shadow model is used to generate member and non-member prediction outputs, and a binary attack classifier learns the corresponding membership signal from these responses. We retain this attack formulation across all experimental conditions so that changes in attack performance reflect changes in the information exposed by the trained SNN rather than differences in the attack methodology. MIA therefore serves as a consistent analytical measure for examining how temporal representation, stochasticity, neuron dynamics, and learning mechanisms affect membership leakage.
\section{Experimental Methodology}

\subsection{\textbf{Datasets and Model Architecture}}

We evaluate the temporal dynamics across five classification benchmarks:
\textit{MNIST}~\cite{lecun1998gradient},
\textit{Fashion-MNIST (FMNIST)}~\cite{xiao2017fashion},
\textit{CIFAR-10}~\cite{krizhevsky2009learning},
\textit{Iris}~\cite{fisher1936use}, and
\textit{Breast Cancer Wisconsin (Diagnostic)}~\cite{street1993nuclear}.
The first three datasets represent image-classification tasks, while Iris and
Breast Cancer provide lower-dimensional tabular settings. MNIST contains
$60{,}000$ training and $10{,}000$ test images of handwritten digits represented
as $28\times28$ grayscale images across 10 classes. Fashion-MNIST follows the
same image dimensions and train/test sizes but contains clothing images from
10 categories. CIFAR-10 contains $50{,}000$ training and $10{,}000$ test
$32\times32$ RGB images across 10 object classes. Iris contains 150 samples
described by four numerical features across three classes, while the Breast
Cancer dataset contains 569 samples represented by 30 numerical features for
binary classification.

For CIFAR-10, training images are augmented using random $32\times32$ crops
with four-pixel padding and random horizontal flipping. Both training and test
images are converted to tensors and normalized using a mean and standard
deviation of $0.5$ for each RGB channel. MNIST and FMNIST images are resized to
$28\times28$, converted to grayscale tensors, and processed consistently across
their training and test partitions. For Iris and Breast Cancer, the data are
divided into training and test sets using an $80/20$ split with a fixed random
state of 42. Numerical features are standardized using a
\texttt{StandardScaler} fitted on the corresponding training set and subsequently
applied to the test set.

For MNIST and FMNIST, a convolutional SNN with two convolutional layers is used.
The layers contain 32 and 64 output channels, respectively, with $3\times3$
kernels and max-pooling after each convolutional stage. The resulting
representation is flattened and passed through a fully connected layer with
1000 hidden units followed by a LIF layer. A final fully connected layer maps
the representation to 10 output classes and is followed by another LIF layer.

For CIFAR-10, a larger convolutional SNN is used to accommodate the greater
complexity of natural RGB images. The network contains four convolutional
layers with 64, 64, 128, and 128 output channels. Batch normalization is
applied after each convolutional layer, while max-pooling is applied after the
second and fourth convolutional layers. The resulting representation is passed
through fully connected layers with 1024 and 512 units, each followed by LIF
dynamics, before reaching the 10-unit LIF output layer.

For the tabular datasets, a fully connected SNN is used. The Iris model maps
four input features to a hidden layer with 1000 units followed by a three-unit
output layer. The Breast Cancer model follows the same structure with 30 input
features and a two-unit output layer. LIF neurons are used after both fully
connected layers. Across all models, the membrane decay parameter is fixed at
$\beta=0.95$.

The model architecture and neuron parameters remain fixed across temporal
dynamics conditions within each dataset. Only the temporal dynamics or the
experimental factor under investigation are changed. This controlled setup
allows differences in classification performance and membership-inference
leakage to be examined without introducing changes in model capacity.

\subsection{\textbf{Biologically Inspired Temporal Dynamics}}

\begin{figure*}[t]
    \centering
    \includegraphics[
        width=\linewidth,
        trim={0 37mm 0 12mm},
        clip
    ]{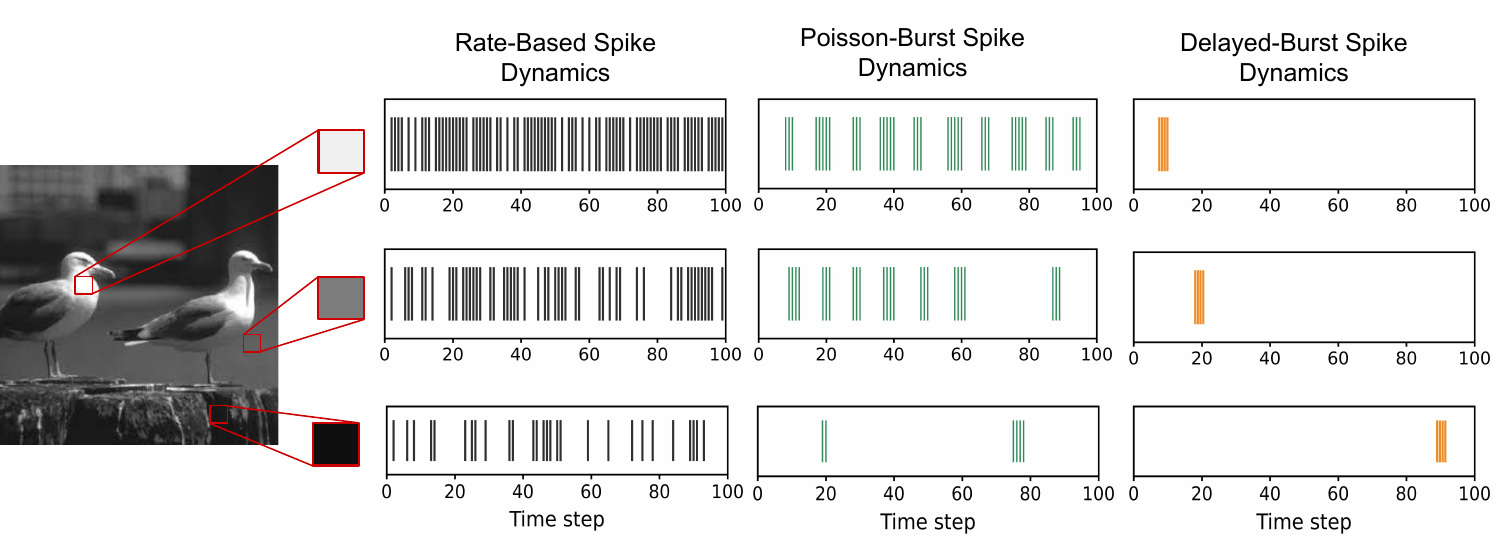}
    \caption{
    Illustration of the three temporal dynamics schemes used in this study.
    Three image regions with high, intermediate, and low normalized input
    intensity are shown from top to bottom. Rate dynamics represents input
    magnitude through spike frequency, such that stronger inputs generate
    denser spike trains. Poisson-Burst dynamics generates stochastic bursts
    at discrete temporal intervals, where stronger inputs increase the
    probability of burst occurrence and the number of spikes within each
    triggered burst is sampled from a Poisson distribution. Delayed-Burst
    dynamics represents input magnitude primarily through burst latency,
    where stronger inputs produce earlier burst onset and weaker inputs
    produce later responses. Each Delayed-Burst event contains a fixed
    number of spikes with a fixed inter-spike interval.
    }
    \label{fig:temporal_spike_generation}
\end{figure*}

To investigate the influence of temporal firing behavior on privacy, we employ
three temporal dynamics schemes: \textit{Rate}, \textit{Poisson Burst}, and
\textit{Delayed Burst}. The two burst-based schemes are adopted from our
previous work~\cite{moshruba2025izhikevich}, where they were introduced as
scalable mechanisms for approximating characteristic temporal behaviors
associated with more biologically expressive neuron models. Rather than
replacing the underlying LIF neurons with the computationally more complex
Izhikevich formulation, the temporal behavior is introduced during spike
encoding while the downstream LIF architecture is kept unchanged. This
separation allows us to study the effect of temporal representation while
controlling for the underlying neuron model. Figure~\ref{fig:temporal_spike_generation}
illustrates how the three dynamics schemes respond to different input
intensities.

\textbf{Rate Dynamics:}
Rate dynamics serves as the baseline temporal representation. Each normalized
input value determines the probability of generating a spike at each simulation
step. Higher input intensities therefore produce denser spike trains, while
lower intensities produce progressively sparser activity, as illustrated in
the first column of Figure~\ref{fig:temporal_spike_generation}. All inputs are
normalized to the range $[0,1]$ before temporal dynamics.

\textbf{Poisson Burst Dynamics:}
Poisson Burst dynamics introduces stochastic burst activity. The simulation
window is divided into fixed temporal intervals, and a Bernoulli trial is
performed at each interval for every input feature. The probability of
initiating a burst is determined by the normalized input intensity, such that
stronger inputs are more likely to generate bursts. When a burst occurs, its
size is sampled from a Poisson distribution and constrained to a predefined
range, while spikes within the burst are separated by a fixed inter-spike
interval. In our implementation, the Poisson mean is set to
$\lambda=3.0$, candidate burst events are evaluated every $5$ timesteps,
the sampled burst size is clipped to the range $[1,8]$, and the inter-spike
interval is fixed at $1$ timestep. The first possible burst can occur at
$t=0$. As shown in the middle column of
Figure~\ref{fig:temporal_spike_generation}, stronger inputs consequently
produce more frequent burst events, while the number of spikes within each
burst remains stochastic.

\textbf{Delayed Burst EDynamics:}
Delayed Burst dynamics represents input magnitude primarily through burst
onset latency. For each input feature, the expected delay is inversely related
to its normalized intensity, such that stronger inputs are expected to fire
earlier and weaker inputs later in the simulation window. The expected delay
is converted into the parameter of a geometric distribution from which the
actual onset delay is sampled. To limit excessive temporal variance, the
sampled delay is clipped to a bounded interval around its expected value.
In our implementation, the clipping range is controlled by a ratio of
$0.2$ of the maximum allowable delay. Once the burst begins, $4$ spikes are
generated with an inter-spike interval of $1$ timestep. The gain and offset
parameters are set to $1.0$ and $0.0$, respectively, and the earliest possible
burst onset is $t=0$. This behavior is illustrated in the final column of
Figure~\ref{fig:temporal_spike_generation}, where stronger inputs produce
earlier bursts and weaker inputs produce progressively later responses.

\subsection{\textbf{Learning Algorithms}}
\label{sec:learning-algorithms}
\subsubsection{\textbf{Backpropagation Through Time (BPTT):}}

Gradient-based optimization was performed using BPTT, which enables temporal credit assignment across the simulation horizon. Since spike generation is inherently non-differentiable, the backward pass employed the Fast Sigmoid surrogate gradient to approximate the derivative of the spiking function, allowing gradients to propagate through the membrane dynamics while preserving discrete spike generation during the forward simulation. This approach enables end-to-end optimization of the network without altering the underlying neuron model.

During training, each input sample was first converted into a temporal spike sequence using one of the three dynamics schemes described in Section3.2. The encoded spike trains were then propagated through the network over $T=10$ discrete simulation steps, where membrane potentials evolved according to the LIF neuron dynamics. Output spike counts accumulated across the simulation window were used to compute the cross-entropy loss, after which gradients were propagated through both the spatial network layers and temporal simulation steps using BPTT.

All networks were optimized using the Adam optimizer with a learning rate of $1\times10^{-3}$ and a batch size of 128 for 50 training epochs. Unless otherwise specified, the membrane decay constant was fixed at $\beta=0.95$, the membrane threshold at $V_{th}=0.2$, and identical network architectures and optimization settings were maintained across all temporal dynamics. Consequently, any observed differences in predictive performance or privacy can be attributed to the temporal dynamics strategy rather than variations in the optimization procedure.

\subsubsection{\textbf{Evolutionary Learning:}}

To examine whether the observed privacy characteristics extend beyond gradient-based optimization, we additionally trained SNNs using an evolutionary learning framework. Unlike BPTT, evolutionary optimization directly searches the parameter space without relying on gradient information, making it naturally applicable to spiking networks with non-differentiable neuron dynamics.

Evolutionary optimization was implemented using the EONS framework. Each individual in the population represents a candidate SNN whose synaptic parameters constitute its genome. During every generation, all candidate networks were evaluated on the complete training set, and classification accuracy was used as the fitness objective. Based on their fitness values, parent networks were selected according to the EONS selection strategy and used to generate offspring through crossover and mutation. The newly generated offspring formed the next population while elite individuals were preserved to retain the best-performing solutions across successive generations. This evolutionary process was repeated until the predefined termination criterion was satisfied.

Due to the substantial computational cost associated with repeatedly evaluating an entire population of candidate networks, evolutionary learning was performed only on the Iris and Breast Cancer datasets. These datasets provide a computationally tractable setting while still enabling meaningful comparisons between learning paradigms. Throughout all evolutionary experiments, the temporal dynamics schemes remained identical to those used in the gradient-based experiments, ensuring that differences in performance or privacy arise solely from the learning algorithm rather than the temporal representation.

\subsection{\textbf{Attack and Evaluation Metrics}}
\label{sec:mia}

\begin{figure*}[t]
    \centering
    \includegraphics[
        width=\linewidth,
        trim={0 0mm 0 0mm},
        clip
    ]{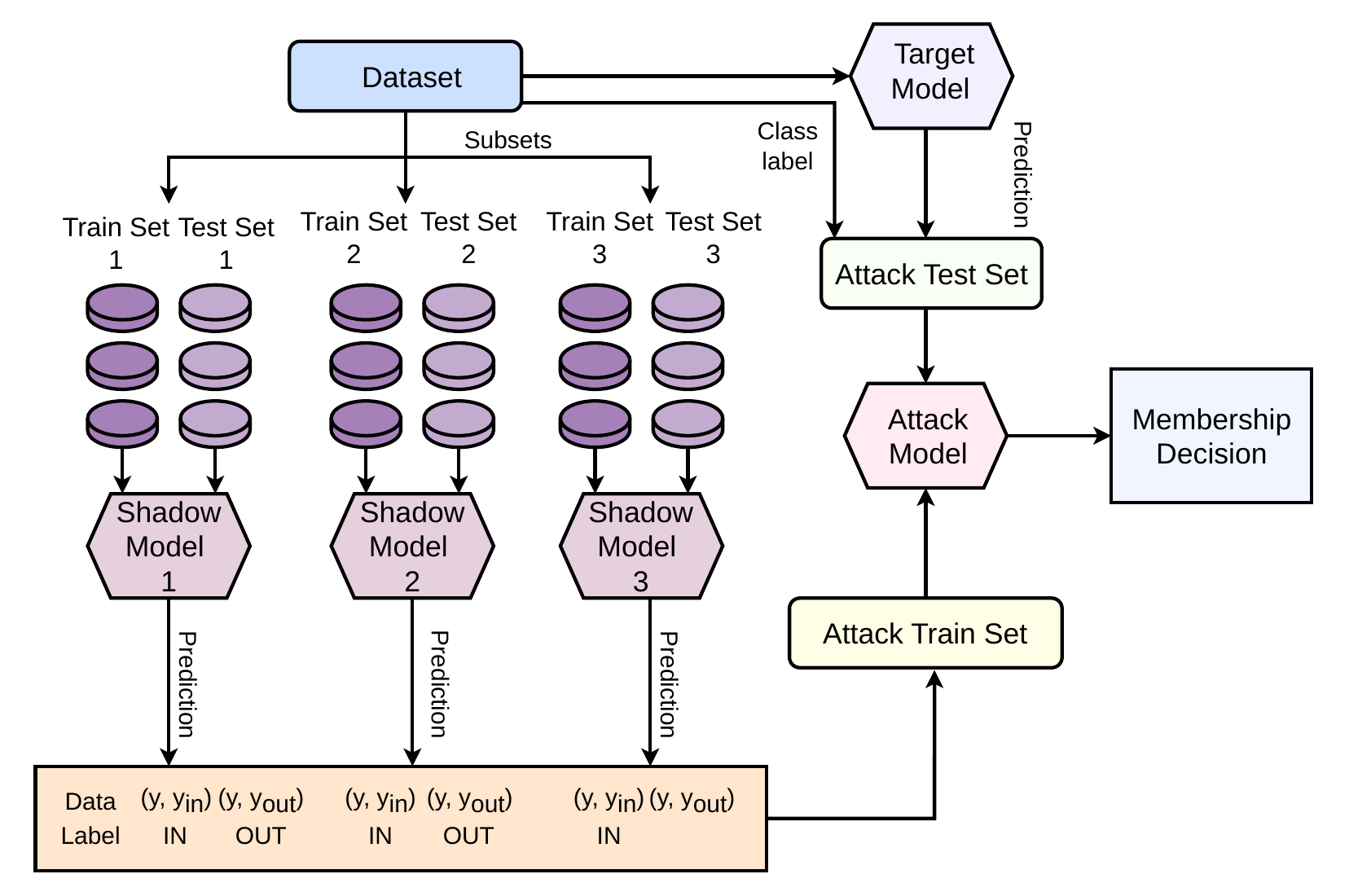}
    \caption{
    MIA framework using three shadow models.
    Three shadow models with the same architecture are independently trained
    using separate shadow train/test splits. For each shadow model, representations
    obtained from its training samples are labeled as members (\textit{IN}), while
    representations obtained from its held-out test samples are labeled as
    non-members (\textit{OUT}). The outputs from all three shadow models are pooled
    to construct a larger attack training set. The target model is queried in the
    same manner to construct the attack test set, and the trained attack model
    predicts whether each target sample belongs to the target model's training set.
    }
    \label{fig:mia_fig}
\end{figure*}
Privacy leakage is evaluated using a black-box MIA, which determines whether a given sample belongs to the target model's
training set from the model representation produced for that sample.
The same attack architecture and evaluation procedure are maintained across
all experiments so that changes in attack performance can be attributed to
the temporal dynamics, neuron models, or learning mechanisms being evaluated.

The attack follows the shadow-model framework illustrated in
Figure~\ref{fig:mia_fig}. To construct a larger and more diverse attack training
set, three shadow models with the same architecture as the target model are
trained independently using different shadow train/test splits. Samples used
to train a shadow model are treated as members (\textit{IN}), while samples
from its corresponding held-out split are treated as non-members
(\textit{OUT}). Each sample is queried once, following the original attack
setting. The final membrane-potential representation at the last simulation
timestep is extracted as the attack feature. The member and non-member
representations produced by all three shadow models are then pooled into a
single attack training set.

For attack evaluation, the target model is queried using its training and test
samples. Representations from the target training set are labeled as members,
while representations from the target test set are labeled as non-members,
which together form the attack test set shown in Figure~\ref{fig:mia_fig}.
Attack examples are grouped according to their class labels, and a class-specific
radial basis function (RBF) support vector machine is trained using the pooled
shadow-model representations. The trained attack classifier is then applied to
the corresponding target-model representations to infer membership. The use of
multiple shadow models therefore increases the amount and diversity of attack
training data without changing the membership decision procedure itself.

Privacy leakage is quantified using the Area Under the Receiver Operating
Characteristic Curve (AUC). An AUC of $0.5$ indicates performance equivalent
to random guessing, while values approaching $1.0$ indicate increasingly
successful membership inference. Therefore, AUC values closer to $0.5$
represent lower observable membership leakage under the evaluated attack.

Model utility is measured using classification accuracy on the corresponding
test set. Together, test accuracy and MIA AUC characterize the
privacy-performance tradeoff across the different temporal dynamics, neuron
models, and learning mechanisms evaluated in this study.
\section{Characterizing Privacy in Biologically Inspired Temporal Dynamics}

\subsection{\textbf{Homogeneous Temporal Dynamics}}
\label{sec:homogeneous}

We first establish a homogeneous baseline by evaluating each temporal dynamic independently. In this setting, all input features within a given experiment follow the same temporal behavior: \textit{Rate}, \textit{Poisson-Burst}, or \textit{Delayed-Burst}. The underlying model architecture, LIF neuron parameters, training procedure, and evaluation protocol remain unchanged across the three conditions. This comparison isolates the effect of temporal firing behavior and provides a direct baseline for examining how each dynamic influences classification performance and membership-inference leakage before introducing heterogeneous combinations.

\subsubsection{\textbf{Results and Analysis}}

\begin{figure*}[t]
    \centering
    \includegraphics[
        width=\linewidth,
        trim={0 27mm 0 27mm},
        clip
    ]{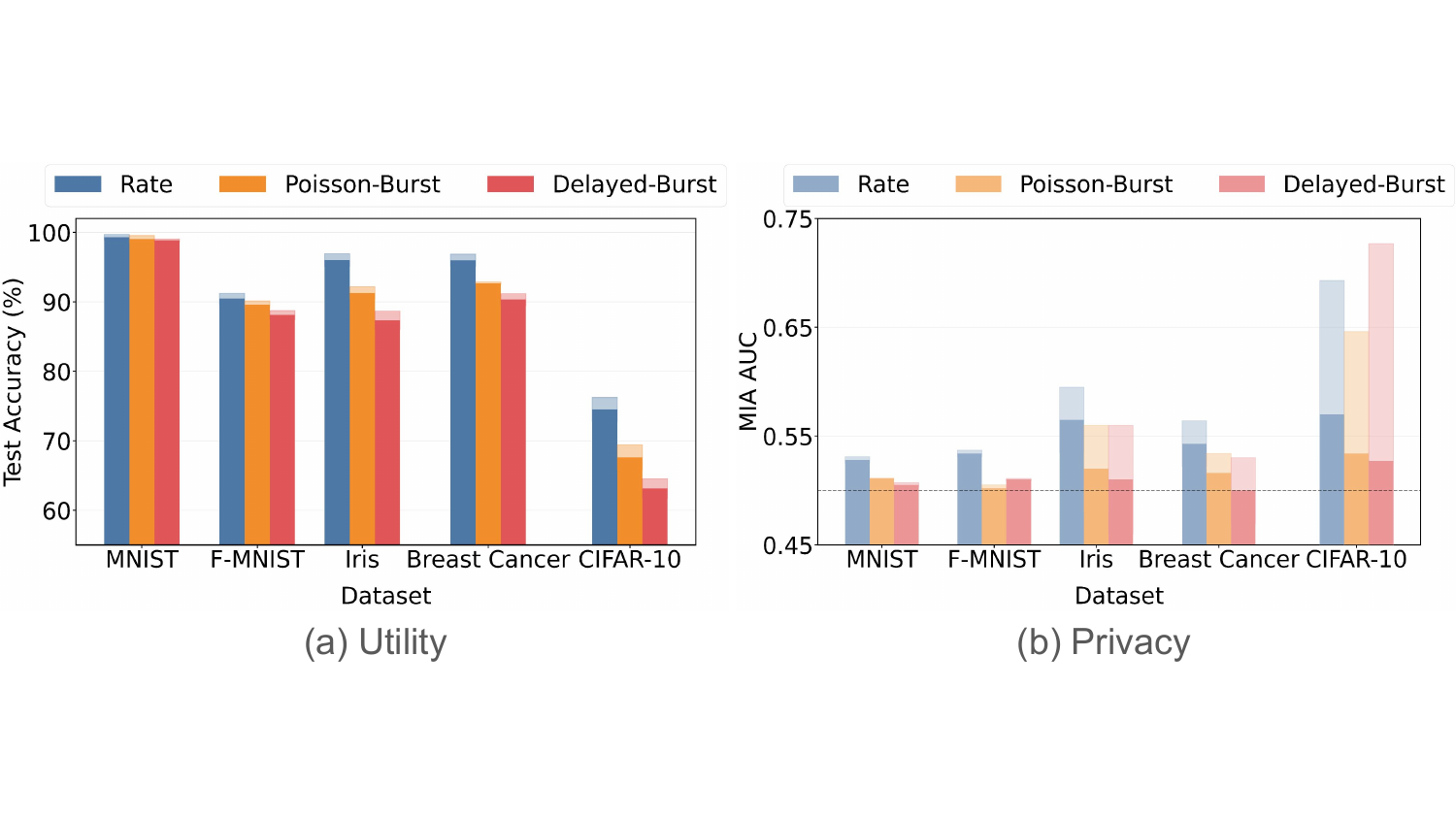}
    \caption{(a) Utility and (b) Privacy under homogeneous temporal dynamics. Bars show the three-run mean, and shaded regions indicate $\pm 1$ standard deviation.}
    \label{fig:hom}
\end{figure*}

Figure~\ref{fig:hom}(a) establishes the homogeneous temporal dynamic baseline used throughout this study. Rate  consistently provides the highest utility, while Poisson-Burst introduces a moderate reduction and Delayed-Burst generally produces the largest degradation. For example, on Breast Cancer, Delayed-Burst results in a $5.89\%$ relative accuracy degradation from the Rate baseline, while on CIFAR-10 the degradation reaches $15.28\%$. This trend suggests that Rate dynamics preserves input information more directly, whereas the clustered and delayed spike activity introduced by the burst-based dynamics makes temporal integration increasingly difficult for the downstream LIF network.

Figure~\ref{fig:hom}(b) shows the corresponding privacy baseline. We report the MIA AUC, defined as $\max(\mathrm{AUC},1-\mathrm{AUC})$, where $0.5$ represents random membership inference and larger values indicate greater leakage. Poisson-Burst and Delayed-Burst generally reduce membership leakage relative to Rate dynamics. For example, on F-MNIST, Poisson-Burst reduces the MIA AUC from $0.534$ to $0.502$, while on Breast Cancer, Delayed-Burst reduces it from $0.543$ to $0.500$. These results indicate that stronger temporal structure can reduce the separability of member and non-member representations, although the larger variability observed on CIFAR-10 suggests that this effect becomes less stable for more complex inputs.

The homogeneous results establish a consistent privacy--utility tradeoff across the temporal dynamics. Rate preserves classification performance most effectively, while Poisson-Burst and Delayed-Burst generally reduce membership leakage at the cost of lower accuracy. The magnitude of this tradeoff varies across datasets, but the overall pattern shows that temporal firing behavior alone is sufficient to alter both predictive performance and membership leakage under an otherwise fixed LIF configuration.
\subsection{\textbf{Heterogeneous Temporal Dynamics}}

Biological neural systems exhibit substantial diversity in their firing behavior. Neurons within the same neural population can show different response patterns, including regular spiking, fast spiking, bursting, and adaptation \cite{connors1990intrinsic,gouwens2019classification}. This diversity has also motivated heterogeneous SNNs, where different neurons are assigned distinct dynamical properties rather than assuming a uniform neuronal population \cite{pereznieves2021neural}. Motivated by this observation, we ask whether introducing analogous diversity into the temporal representation of the input affects the privacy-performance characteristics of SNNs. We do not attempt to reproduce biological neuronal heterogeneity directly. Instead, we construct a simplified heterogeneous temporal dynamics that combines the Rate, Poisson-Burst, and Delayed-Burst dynamics examined in the homogeneous setting.

Let an input sample contain $D$ dynamics dimensions. Each input dimension $d$ is assigned one fixed temporal dynamic,
\begin{equation}
    z_d \in \{\mathrm{Rate}, \mathrm{Poisson\_Burst}, \mathrm{Delayed\_Burst}\}.
\end{equation}
This formulation follows the general idea that heterogeneity exists across units within a population rather than assigning a common dynamical behavior to the entire population \cite{pereznieves2021neural}. The assignment remains fixed throughout training, testing, and membership-inference evaluation, while the spike activity generated by each dimension continues to depend on its input value. Consequently, a single sample can contain Rate, Poisson-Burst, and Delayed-Burst activity across different input dimensions. The population composition is controlled through user-defined proportions
\begin{equation}
    [\pi_{\mathrm{R}}, \pi_{\mathrm{PB}}, \pi_{\mathrm{DB}}],
\end{equation}
where the proportions specify the fraction of input dimensions assigned to each temporal dynamic. We consider both two-way and three-way heterogeneous configurations. In the two-way settings, $80\%$ of the input dimensions use Rate dynamics, while the remaining $20\%$ use either Poisson-Burst (\textit{80R/20PB}) or Delayed-Burst (\textit{80R/20DB}). We also consider an \textit{80R/10PB/10DB} configuration, where $80\%$ of the input dimensions are assigned to Rate, $10\%$ to Poisson-Burst, and $10\%$ to Delayed-Burst.

For the experiments, heterogeneity is introduced at the same input temporal dynamics stage used for the homogeneous configurations. For image datasets, each image is first flattened into its input dimensions and normalized before temporal spike generation. Rate, Poisson-Burst, and Delayed-Burst spike representations are generated directly from the corresponding input values using the same parameter settings as in their homogeneous counterparts. A fixed input-level assignment then selects the appropriate temporal representation for each dimension. The resulting heterogeneous spike tensor is reshaped to the original spatial dimensions and provided to the convolutional SNN, such that all configurations share the same downstream network architecture and differ only in the temporal organization of the input spike representation.

We evaluate different mixture ratios against the homogeneous Rate baselines. Each configuration is repeated across three random seeds. For each seed, a new fixed input-level temporal assignment is generated and remains unchanged throughout the corresponding experiment. The target model and shadow models use the same temporal assignment within a run, while their network parameters are independently initialized. We report the average best target-model test accuracy and the average membership-inference AUC across the three runs.

\subsubsection{\textbf{Results and Analysis}}

\begin{figure*}[t]
    \centering
    \includegraphics[
        width=\linewidth,
        trim={0 0mm 0 0mm},
        clip
    ]{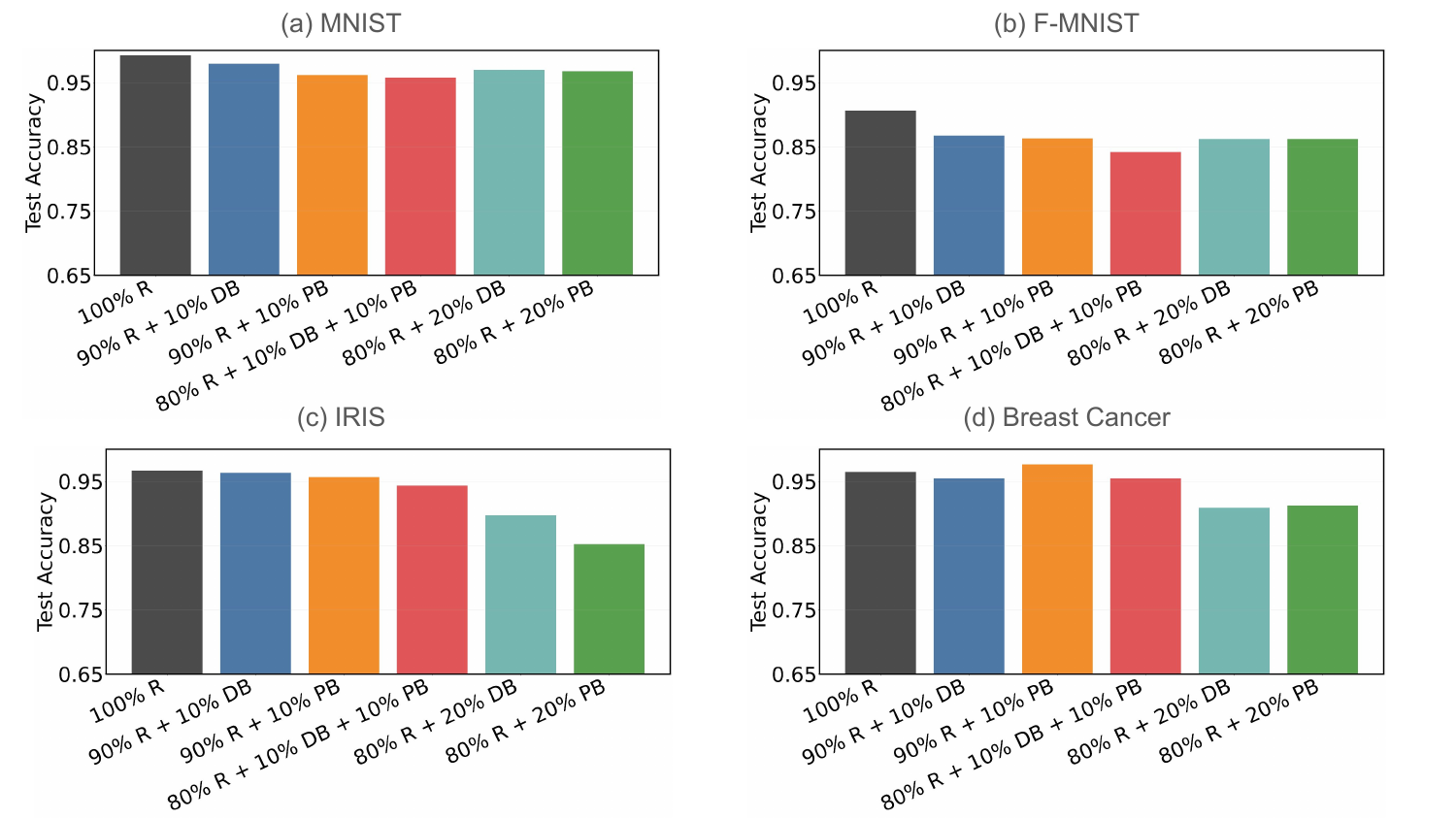}
    \caption{Classification performance under homogeneous Rate dynamics and heterogeneous input-level temporal configurations across MNIST, F-MNIST, Iris, and Breast Cancer. Configuration labels indicate the fraction of input dimensions assigned to each temporal dynamic. For example, \textit{90R/10PB} denotes 90\% Rate and 10\% Poisson-Burst dynamics, while \textit{80R/10PB/10DB} denotes 80\% Rate, 10\% Poisson-Burst, and 10\% Delayed-Burst dynamics.}
    \label{fig:het_perf}
\end{figure*}
The utility results in Fig.~\ref{fig:het_perf} show that heterogeneous temporal dynamics generally introduces a performance cost, although small heterogeneous components are often well tolerated. The degradation becomes more pronounced as a larger fraction of Rate-coded inputs is replaced by PB or DB dynamics. For example, on Iris, the 90R/10DB configuration results in only a $0.34\%$ relative degradation from the homogeneous Rate baseline, whereas the 80R/20PB configuration produces an $11.86\%$ relative degradation. A similar trend is observed across the other datasets, although the magnitude varies with the task. This behavior suggests that a predominantly Rate-coded representation preserves most of the original input information, while increasing the proportion of burst-based dynamics introduces greater temporal variation through clustered or delayed spike activity. The downstream LIF neurons must therefore integrate inputs with increasingly different temporal statistics, which can make information recovery and optimization more difficult.

\begin{figure*}[t]
    \centering
    \includegraphics[
        width=\linewidth,
        trim={0 0mm 0 0mm},
        clip
    ]{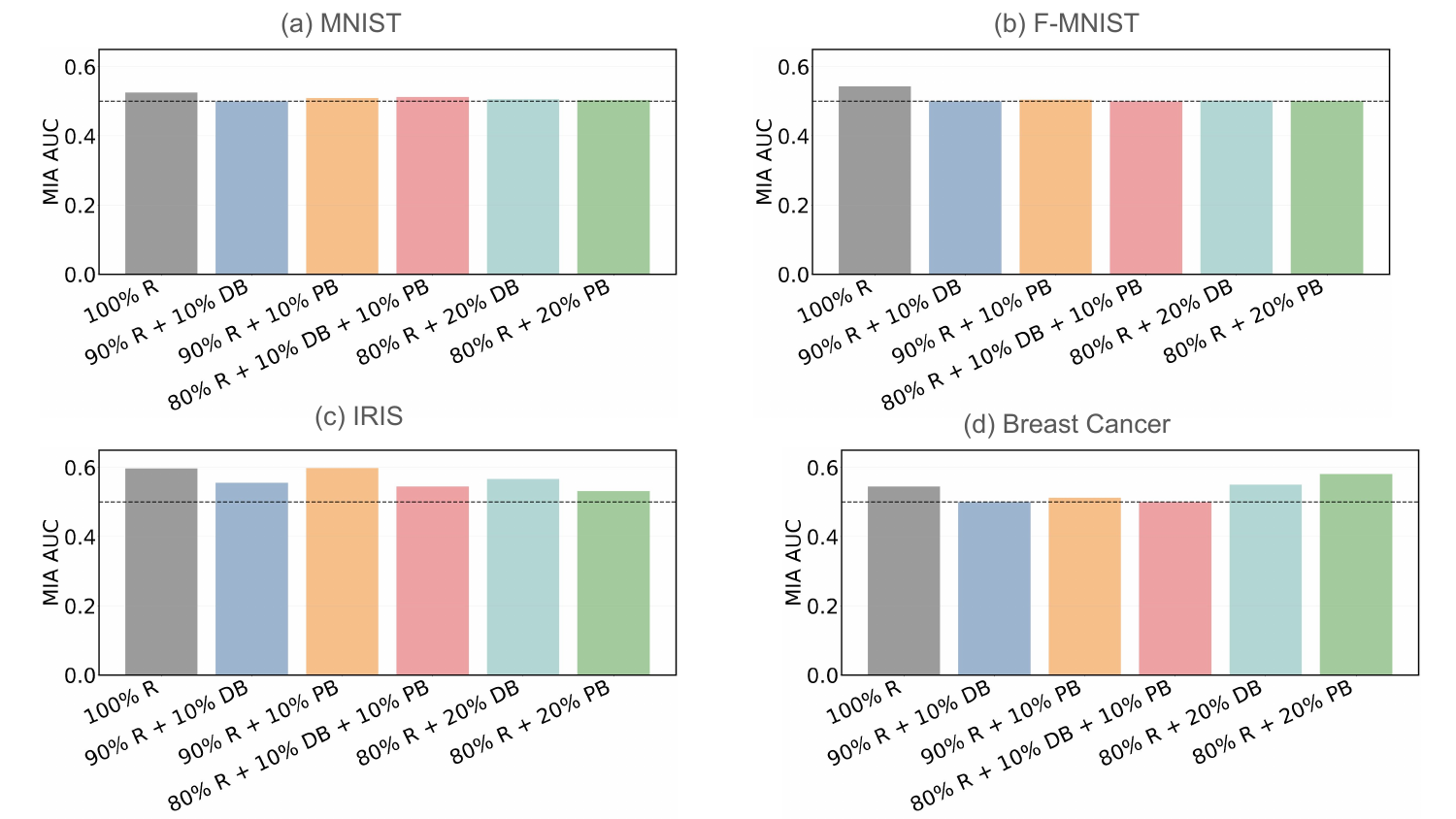}
    \caption{
    Membership-inference vulnerability under homogeneous Rate dynamics and heterogeneous input-level temporal configurations across MNIST, F-MNIST, Iris, and Breast Cancer.}
    \label{fig:het_mia}
\end{figure*}

The privacy results in Fig.~\ref{fig:het_mia} show that heterogeneous temporal dynamics frequently reduces membership leakage, but the improvement is not monotonic with the amount of heterogeneity. We report \textit{direction-invariant MIA AUC}, computed as $\max(\mathrm{AUC},1-\mathrm{AUC})$, so that values below $0.5$ are not incorrectly interpreted as stronger privacy when the attack decision can simply be reversed. Across the datasets, several heterogeneous configurations move attack performance substantially closer to the chance level of $0.5$. For example, on F-MNIST, heterogeneous configurations reduce the MIA AUC from the Rate baseline of $0.543$ to approximately $0.500$--$0.504$. However, increasing the Poisson Burst or Delayed Burst proportion does not consistently produce further privacy gains, which indicates that leakage depends on the specific temporal composition rather than on the amount of heterogeneity alone. A plausible explanation is that moderate temporal diversity disrupts consistent member-specific response patterns by distributing input information across different spike-timing structures, which reduces the separability of member and non-member representations. Excessive heterogeneity, however, can distort the representation sufficiently to reduce utility without necessarily providing additional privacy.

This analysis shows that introducing temporal diversity can reduce membership leakage while preserving much of the classification performance when the burst-based components remain limited. However, increasing the degree of heterogeneity does not lead to progressively stronger privacy, and larger deviations from the Rate-dominant representation can impose substantial utility costs without additional reduction in leakage. The privacy--utility behavior therefore depends on the specific composition of Rate, Poisson-Burst, and Delayed-Burst dynamics rather than on the amount of temporal heterogeneity alone.
\section{Stochasticity and Privacy}
\label{sec:stochasticity}
This section examines whether the privacy differences associated with distinct temporal spike representations can be explained by stochasticity, rather than by temporal structure alone. The objective is to disentangle the contribution of random perturbations from the intrinsic organization of the spike trains and to determine whether reduced membership leakage is primarily a consequence of increased representational variability. To isolate this factor, we introduce controlled perturbations at three levels: temporal jitter, which modifies spike timing; input noise, which perturbs the signal before spike generation; and spike noise, which alters the generated spike trains. The resulting changes in classification performance and membership-inference vulnerability are then compared to assess whether stochasticity provides a consistent explanation for the privacy behavior observed across the temporal-dynamics experiments.

\subsection{\textbf{Temporal Jitter}}

\begin{figure*}[t]
    \centering
    \includegraphics[
        width=\linewidth,
        trim={0 0mm 0 0mm},
        clip
    ]{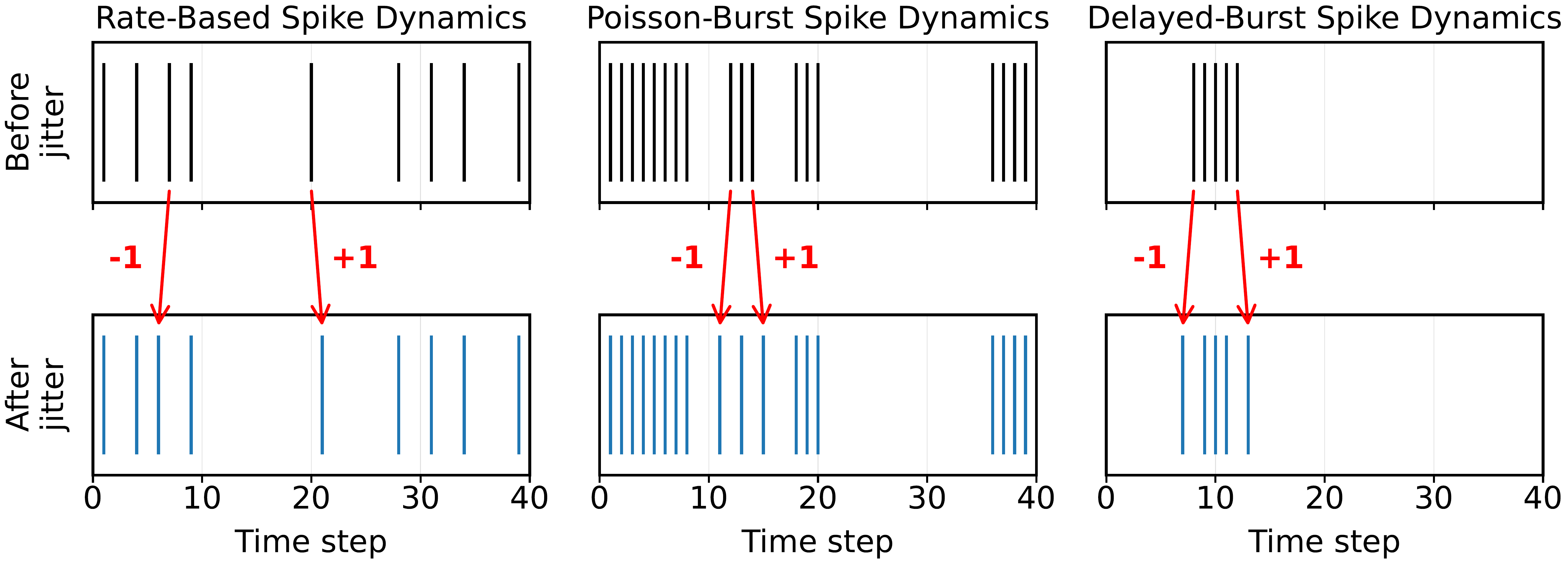}
    \caption{Illustration of temporal jitter applied to Rate, Poisson-Burst, and Delayed-Burst spike trains. Selected spikes are shifted by one timestep ($\pm1$), while the overall spike pattern and spike count are preserved.}
    \label{fig:jitter}
\end{figure*}

Temporal precision constitutes a defining property of spike-based representations and may influence how training-specific information is expressed within an SNN. In the homogeneous experiments, Rate, Poisson-Burst, and Delayed-Burst dynamics exhibited different privacy characteristics despite using the same downstream LIF architecture. One possible explanation is that these differences arise from the degree to which information is encoded in the exact timing of spikes. To examine this possibility, we perturb spike timing while preserving the overall firing pattern and spike count, thereby isolating temporal precision from other properties of the dynamics.

For a spike occurring at timestep $t_i$, temporal jitter produces a perturbed spike time
\begin{equation}
    t_i' = t_i + \delta_i,
    \qquad
    \delta_i \in \{-1,+1\}.
\end{equation}
A selected spike is shifted one timestep earlier or later only when the destination timestep is valid and unoccupied; otherwise, the spike remains at its original location. This constraint preserves the total spike count and prevents multiple spikes from collapsing onto the same timestep. Figure~\ref{fig:jitter} illustrates the resulting perturbation for Rate, Poisson-Burst, and Delayed-Burst spike trains.

We evaluate three perturbation levels by selecting $20\%$, $50\%$, and $100\%$ of spikes for jitter. The same procedure is applied independently to each temporal dynamics, while the downstream network architecture, training protocol, and membership-inference evaluation remain unchanged. This controlled design enables direct assessment of whether reduced temporal precision alters classification performance or membership leakage, and therefore whether exact spike timing is a significant contributor to the privacy behavior observed across temporal dynamics.

\subsubsection{\textbf{Result and Analysis}}

\begin{figure*}[t]
    \centering
    \includegraphics[
        width=\linewidth,
        trim={0 20mm 0 42mm},
        clip
    ]{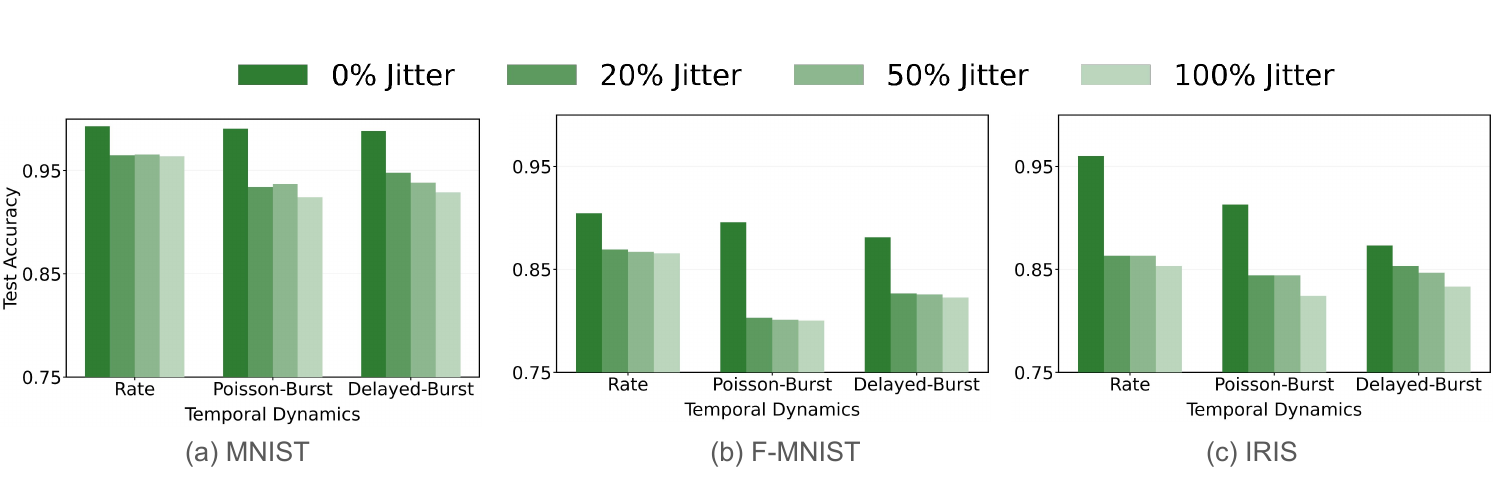}
    \caption{Classification utility under increasing temporal jitter for Rate, Poisson-Burst, and Delayed-Burst dynamics.}
    \label{fig:jitter_perf}
\end{figure*}

Figure~\ref{fig:jitter_perf} shows that perturbing spike timing consistently reduces classification utility, which confirms that temporal precision carries task-relevant information. The degradation generally increases with the amount of jitter and is more pronounced for the burst-based dynamics. For example, on F-MNIST, increasing jitter from $0\%$ to $100\%$ reduces Poisson-Burst accuracy by approximately $10.7\%$ relative to the unperturbed baseline, whereas Rate dynamics decreases by approximately $4.3\%$. This greater sensitivity of the burst-based representations is consistent with their stronger dependence on the temporal organization of clustered spike activity. As progressively more spikes are displaced from their original temporal positions, the structure used by the downstream LIF neurons becomes increasingly disrupted, which reduces the fidelity of the encoded representation.

\begin{figure*}[t]
    \centering
    \includegraphics[
        width=\linewidth,
        trim={0 20mm 0 42mm},
        clip
    ]{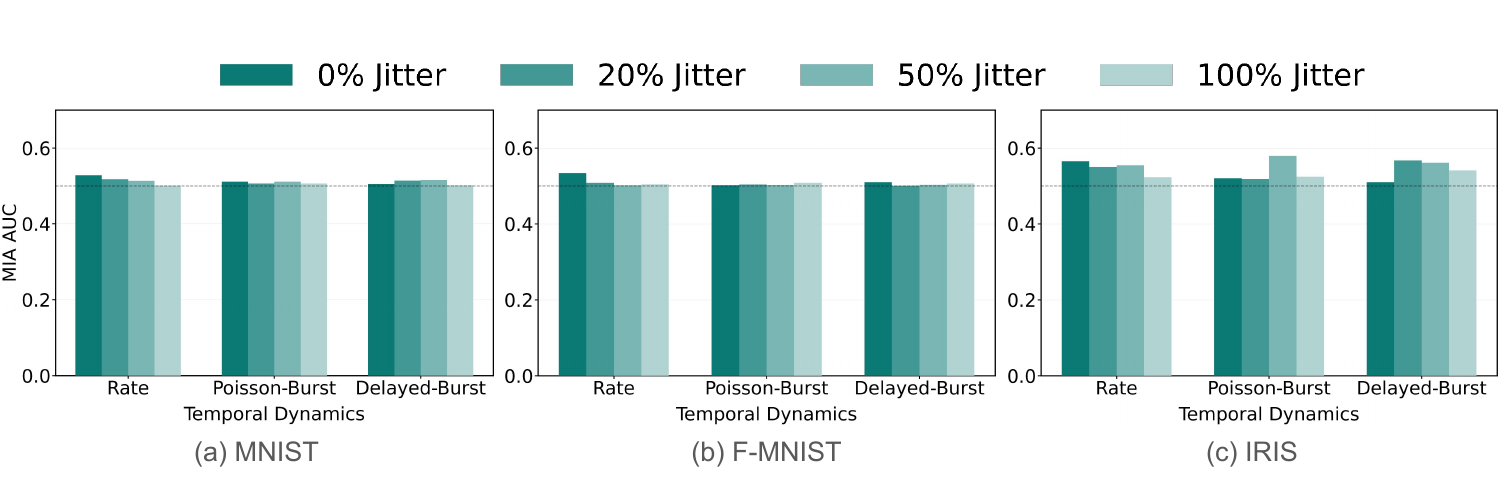}
    \caption{Membership-inference vulnerability under increasing temporal jitter for Rate, Poisson-Burst, and Delayed-Burst dynamics.}
    \label{fig:jitter_mia}
\end{figure*}

Figure~\ref{fig:jitter_mia} examines whether the same increase in temporal stochasticity produces a systematic privacy benefit. Jitter reduces leakage in several configurations; for example, the F-MNIST Rate condition decreases from $0.534$ without jitter to approximately $0.502$ at $50\%$ jitter, a reduction of about $6.0\%$. However, the privacy response is not monotonic across either jitter levels or temporal dynamics, and some configurations exhibit little improvement or increased leakage as jitter increases. Thus, greater temporal variability does not consistently reproduce the privacy differences observed between the original Rate, Poisson-Burst, and Delayed-Burst representations. These results indicate that stochastic perturbation of spike timing can influence membership leakage, but stochasticity alone is insufficient to explain the privacy behavior of the different temporal dynamics.

Thus the temporal-jitter results show that disrupting spike timing consistently weakens classification performance, but the corresponding effect on membership leakage remains irregular across jitter levels and temporal dynamics. This mismatch indicates that temporal precision is important for preserving task-relevant information, while increased timing randomness does not yield a reliable privacy benefit. The privacy differences among Rate, Poisson-Burst, and Delayed-Burst dynamics therefore cannot be explained by temporal stochasticity alone and instead depend on the organization of spike activity over time.

\subsection{\textbf{Input Noise}}

Input noise is introduced to examine whether the privacy differences observed across temporal spike representations can be explained by stochastic variation in the signal before temporal dynamics takes place. If increased randomness alone is responsible for reducing membership leakage, then perturbing the input should produce a consistent privacy benefit even before Rate, Poisson-Burst, or Delayed-Burst dynamics are generated. This experiment therefore tests whether stochasticity at the input level can account for the privacy behavior observed across temporal dynamics, or whether the structure introduced during spike generation remains necessary to explain those differences.

To isolate this effect, Gaussian noise is added directly to the normalized input before spike generation. For an input value $x \in [0,1]$, the perturbed input is defined as

\begin{equation}
    x' = \mathrm{clip}(x + \epsilon, 0, 1),
    \qquad
    \epsilon \sim \mathcal{N}(0,\sigma^2),
\end{equation}

where $\sigma$ controls the noise magnitude. We evaluate three noise levels, $\sigma \in \{0.10, 0.25, 0.50\}$, corresponding to progressively stronger perturbations of the input signal. The perturbed input is then converted into Rate, Poisson-Burst, or Delayed-Burst spike trains using the same dynamics procedure as in the baseline experiments.

\subsubsection{\textbf{Result and Analysis}}

\begin{figure*}[t]
    \centering
    \includegraphics[
        width=\linewidth,
        trim={0 32mm 0 35mm},
        clip
    ]{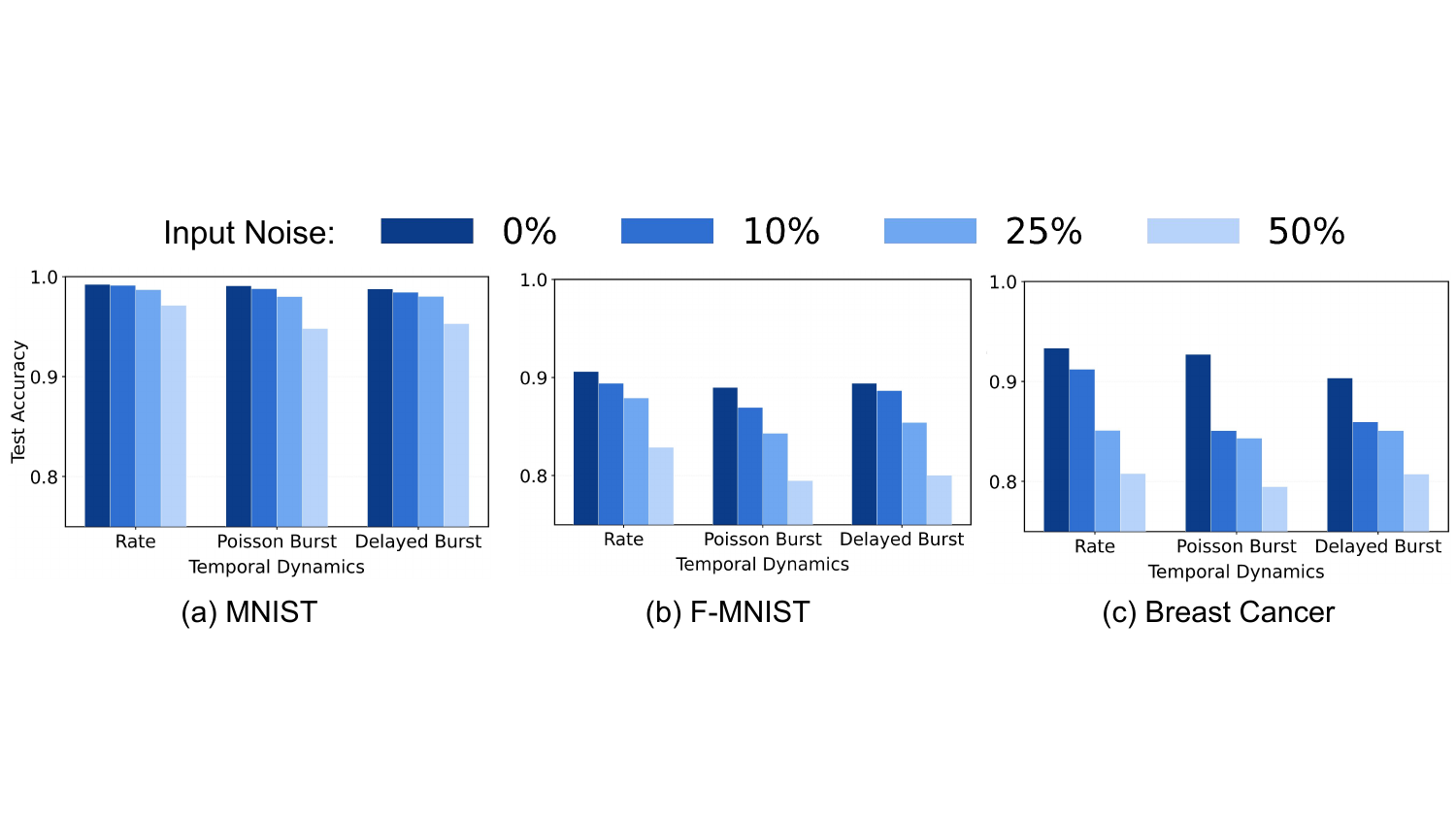}
    \caption{Classification utility under increasing input noise for Rate, Poisson-Burst, and Delayed-Burst dynamics on MNIST, F-MNIST, and Breast Cancer.}
    \label{fig:inp_noise_perf}
\end{figure*}

Figure~\ref{fig:inp_noise_perf} shows a consistent decline in classification performance as input noise increases. The effect is modest on MNIST at lower noise levels but becomes substantially stronger on F-MNIST and Breast Cancer, particularly under the burst-based dynamics. For example, Poisson-Burst on F-MNIST exhibits approximately $12.1\%$ relative accuracy degradation at $50\%$ input noise. This trend indicates that stronger perturbation before spike generation progressively removes task-relevant information from which the temporal representation is formed.

\begin{figure*}[t]
    \centering
    \includegraphics[
        width=\linewidth,
        trim={0 30mm 0 35mm},
        clip
    ]{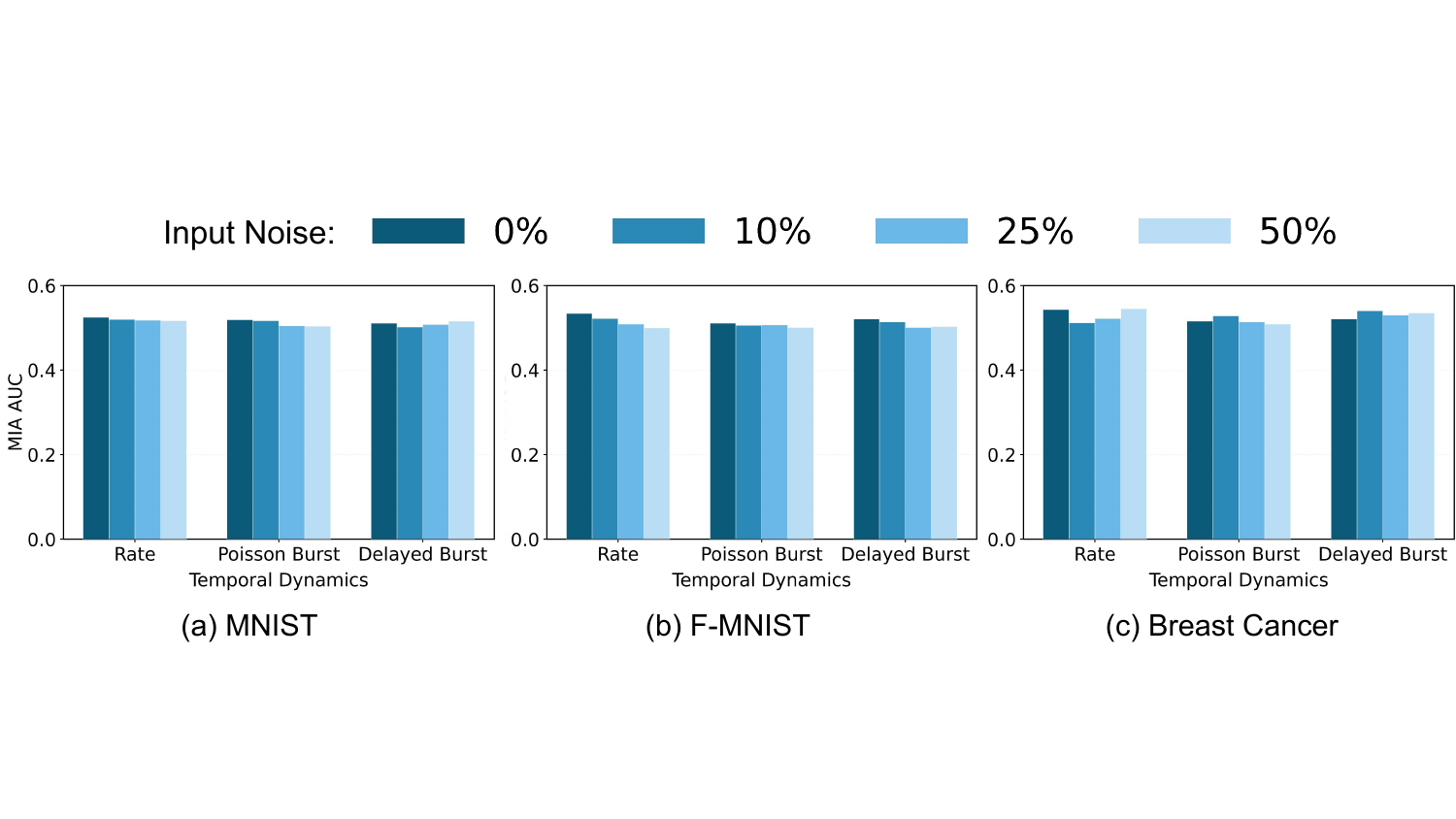}
    \caption{Membership-inference vulnerability under increasing input noise for Rate, Poisson-Burst, and Delayed-Burst dynamics on MNIST, F-MNIST, and Breast Cancer.}
    \label{fig:inp_noise_mia}
\end{figure*}

The privacy response in Figure~\ref{fig:inp_noise_mia} is considerably less systematic. Increasing input noise generally moves membership leakage downward, particularly on F-MNIST, but the reduction is small or non-monotonic for several configurations. Breast Cancer shows the same broader inconsistency, with leakage varying across noise levels and temporal dynamics rather than following a uniform downward trend. More importantly, the three temporal dynamics do not converge toward a common privacy behavior as stochasticity increases. Their responses remain dependent on both the temporal dynamic and the dataset.

Therefore, the input-noise results do not support stochasticity as the primary source of the privacy differences among Rate, Poisson-Burst, and Delayed-Burst dynamics. As the noise level increases, classification performance declines substantially, while membership leakage shows only small, inconsistent, or non-monotonic changes. If increased randomness were the main mechanism responsible for the lower leakage of the burst-based dynamics, stronger perturbation would be expected to produce a clearer privacy benefit without a disproportionate loss in utility. Instead, the dominant effect is utility degradation with no corresponding and consistent gain in privacy, which indicates that stochasticity alone cannot explain the observed privacy behavior.

\subsection{\textbf{Spike Noise}}
Spike noise is used to test whether privacy is influenced by stochastic corruption after the temporal representation has already been formed. Unlike input noise, which perturbs the signal before encoding, this experiment directly alters the generated spike trains while preserving the underlying Rate, Poisson-Burst, or Delayed-Burst dynamics process. This allows us to examine whether randomness at the spike-event level can account for changes in membership leakage, or whether the original temporal organization continues to play an important role after stochastic perturbation is introduced.

For each generated spike train containing $N$ spikes, the number of perturbed spike events is determined by

\begin{equation}
    k = \mathrm{round}(pN),
\end{equation}

where $p \in \{0.10, 0.25, 0.50\}$ denotes the spike-noise level. For each spike train, one of two perturbation operations is selected with equal probability. In the removal condition, $k$ existing spikes are randomly deleted. In the addition condition, $k$ spikes are inserted at randomly selected unoccupied timesteps. A single spike train therefore undergoes either spike addition or spike removal, but not both, and the resulting representation remains binary.

The perturbation is applied after temporal spike generation and before the spike train is passed to the downstream LIF network. All remaining components of the training and three-shadow membership-inference pipeline are kept unchanged. By varying the fraction of spike events affected while keeping the addition-removal probability fixed, this experiment isolates the effect of stochastic spike-event corruption from the temporal dynamics mechanism itself.

\subsubsection{\textbf{Result and Analysis}}

\begin{figure*}[t]
    \centering
    \includegraphics[
        width=\linewidth,
        trim={0 26mm 0 35mm},
        clip
    ]{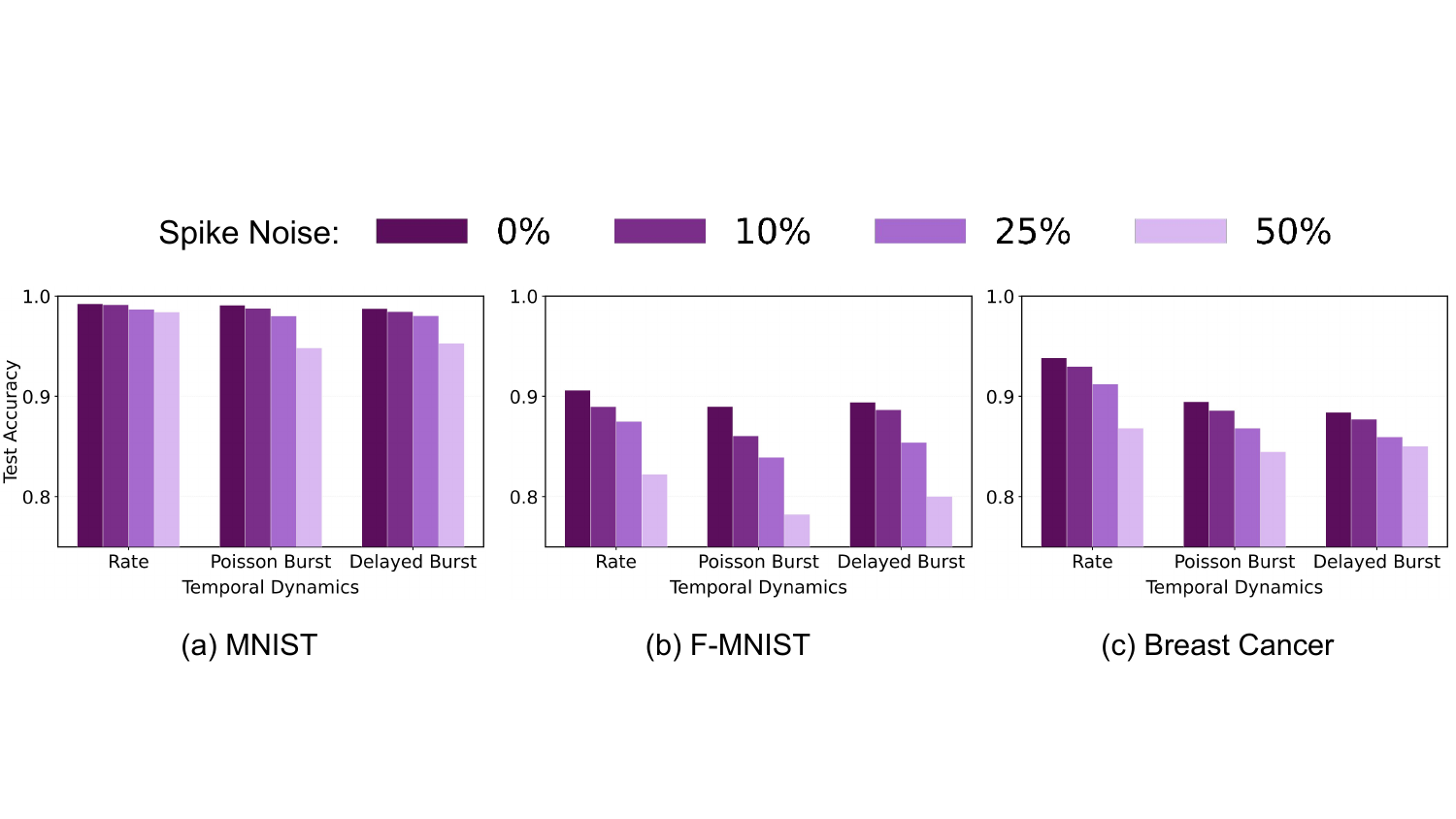}
    \caption{Classification utility under increasing spike noise for Rate, Poisson-Burst, and Delayed-Burst dynamics on MNIST, F-MNIST, and Breast Cancer.}
    \label{fig:spk_noise_perf}
\end{figure*}

\begin{figure*}[t]
    \centering
    \includegraphics[
        width=\linewidth,
        trim={0 30mm 0 35mm},
        clip
    ]{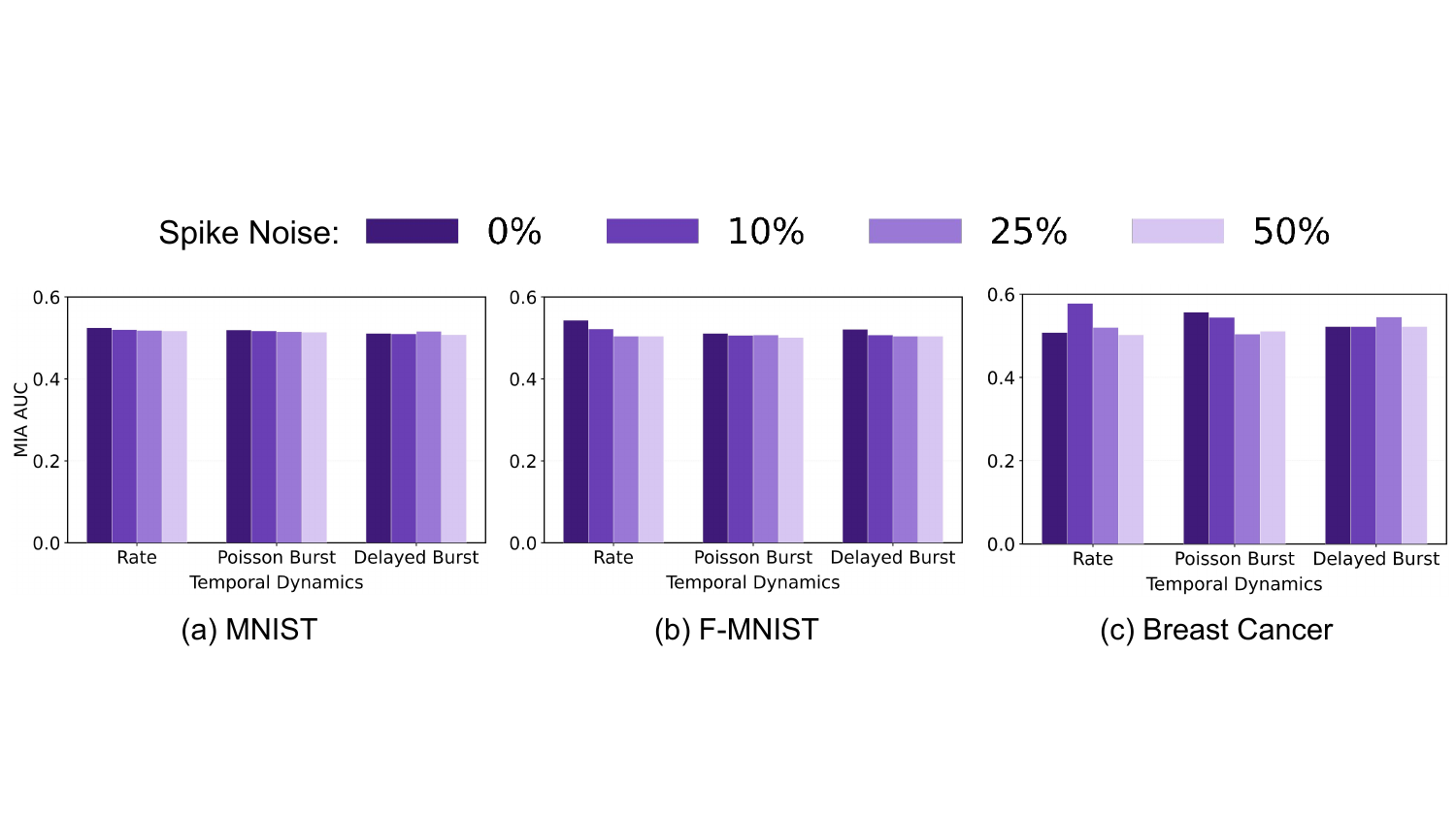}
    \caption{Membership-inference vulnerability under increasing spike noise for Rate, Poisson-Burst, and Delayed-Burst dynamics on MNIST, F-MNIST, and Breast Cancer.}
    \label{fig:spk_noise_mia}
\end{figure*}

Figure~\ref{fig:spk_noise_perf} shows that classification performance remains comparatively stable under low and moderate spike noise, with larger degradation appearing primarily at the highest perturbation level. The effect is more pronounced on F-MNIST and Breast Cancer than on MNIST, where accuracy remains relatively stable across the evaluated noise levels. For example, Poisson-Burst on F-MNIST exhibits approximately $5.1\%$ relative accuracy degradation at $50\%$ spike noise. This behavior suggests that moderate corruption of individual spike events can be tolerated without substantially disrupting task-relevant temporal structure, whereas stronger addition or removal of spikes progressively degrades the information available to the network.

The privacy response in Figure~\ref{fig:spk_noise_mia} shows no consistent relationship between spike-noise intensity and membership leakage. MIA AUC fluctuates across noise levels for all three temporal dynamics, with both increases and decreases appearing depending on the dataset and configuration. This irregularity is particularly evident on Breast Cancer, where several conditions show substantial changes at intermediate noise levels without following a monotonic progression. F-MNIST exhibits a similarly configuration-dependent response, while MNIST remains comparatively stable. These results indicate that perturbing the generated spike trains alters membership leakage, but not in a predictable direction.

This lack of a consistent privacy response contrasts with the clear utility degradation observed under stronger spike noise. Classification performance declines as spike corruption increases, yet this loss of task-relevant information is not accompanied by a corresponding or consistent improvement in privacy. Spike-level stochasticity therefore alters the representation, but it does not reproduce the privacy differences observed among Rate, Poisson-Burst, and Delayed-Burst dynamics. The results instead suggest that these differences depend on the structured temporal organization of the spike activity rather than on random corruption of spike events alone.
\section{Intrinsic Neuron Dynamics and Privacy}

SNNs can be constructed using a range of neuron models that differ in the complexity of their internal dynamics and the firing behaviors they can express. The LIF neuron remains one of the most widely used models because of its computational simplicity and its compatibility with gradient-based SNN training \cite{abbott1999lapicque,wang2022linear}. However, its dynamics are comparatively limited, since the membrane potential primarily integrates incoming current with leakage until a threshold-triggered spike and reset occur. More expressive neuron models can reproduce a substantially broader range of temporal behaviors.

One prominent example is the Izhikevich neuron model, which combines computational efficiency with the ability to generate diverse firing regimes, including regular spiking, bursting, delayed firing, rebound activity, and frequency adaptation \cite{izhikevich2003simple}. These firing characteristics motivated the temporal patterns examined earlier in this work, where Poisson-Burst and Delayed-Burst dynamics were introduced at the input level while the downstream network continued to use LIF neurons. That formulation allowed the temporal representation itself to be studied independently, but it does not address whether the intrinsic dynamics of the neuron model can also influence privacy.
Because intrinsic neuron dynamics shape the evolution of spike activity within the network, they may provide an additional source of variation in privacy behavior beyond the input-level temporal patterns examined earlier \cite{moshruba2024neuromorphic,guan2025privacy}. We therefore investigate whether changing the neuron model itself produces corresponding differences in membership leakage.

Directly replacing the LIF encoding with the Izhikevich formulation is not straightforward within the gradient-based training framework used in this study. We therefore consider the AdEx neuron model, which extends integrate-and-fire dynamics with an exponential membrane term and an adaptation variable \cite{brette2005adaptive}. AdEx can reproduce multiple electrophysiological firing classes while remaining sufficiently compact for integration into trainable SNN architectures \cite{brette2005adaptive}. 

This research therefore shifts the source of temporal diversity from the input encoding to the neuron itself. Rather than externally constructing Rate, Poisson-Burst, or Delayed-Burst spike trains and processing them with the same LIF dynamics, we investigate whether richer firing regimes can emerge from the internal dynamics of AdEx neurons and whether those dynamics produce different privacy--utility behavior. 

\subsection{\textbf{Adaptive Exponential Integrate-and-Fire(AdEx) Neurons}}

The AdEx model extends conventional integrate-and-fire dynamics by combining an exponential spike initiation mechanism with an adaptation variable \cite{brette2005adaptive}. This formulation preserves the relative simplicity of integrate-and-fire neurons while supporting a substantially broader range of firing behaviors. In particular, changes in the adaptation and reset parameters can produce tonic spiking, initial bursting, regular bursting, delayed activity, and other firing regimes \cite{naud2008firing}. These properties make AdEx suitable for examining whether privacy behavior can be influenced by intrinsic neuronal dynamics rather than by input-level temporal structure alone. 

The membrane potential $V$ evolves according to

\begin{equation}
C\frac{dV}{dt}
=
-g_L(V-E_L)
+
g_L\Delta_T
\exp\left(\frac{V-V_T}{\Delta_T}\right)
-w
+
I,
\end{equation}

where $C$ is the membrane capacitance, $g_L$ is the leak conductance, $E_L$ is the resting potential, $V_T$ is the effective threshold, $\Delta_T$ controls the sharpness of spike initiation, $I$ is the input current, and $w$ denotes the adaptation current \cite{brette2005adaptive}. The first term governs passive leakage toward the resting potential, while the exponential term produces the rapid membrane-potential increase associated with spike initiation. As $V$ approaches $V_T$, the exponential contribution becomes increasingly dominant and drives the neuron toward a spike. 

The adaptation variable follows

\begin{equation}
\tau_w\frac{dw}{dt}
=
a(V-E_L)-w,
\end{equation}

where $\tau_w$ determines the adaptation time scale and $a$ controls subthreshold adaptation. Following a spike, the internal state is updated as

\begin{equation}
V \leftarrow V_{\mathrm{reset}},
\qquad
w \leftarrow w+b,
\end{equation}

where $V_{\mathrm{reset}}$ specifies the post-spike membrane potential and $b$ controls spike-triggered adaptation \cite{brette2005adaptive}. The subthreshold adaptation parameter $a$ determines how strongly the adaptation current responds to deviations of the membrane potential from rest, while $\tau_w$ sets the time scale over which this adaptation evolves. The threshold parameters $V_T$ and $\Delta_T$ control the onset and sharpness of the exponential membrane response that precedes spike generation. After a spike, $V_{\mathrm{reset}}$ determines how far the membrane potential is driven from the firing region, whereas $b$ increases the adaptation current and can suppress subsequent firing. Variations in these mechanisms can therefore shift the neuron between regular spiking, adapting, and burst-like regimes \cite{brette2005adaptive,naud2008firing}.

An important property of AdEx is that qualitatively different firing regimes emerge from changes in its intrinsic parameters rather than from externally modifying the input spike train. Naud et al. showed that the same two-equation model can transition among tonic spiking, initial bursting, regular bursting, and adaptation-dominated regimes as its parameter configuration changes \cite{naud2008firing}. This provides a useful contrast to the temporal dynamics considered earlier, where Poisson-Burst and Delayed-Burst patterns were constructed before the input reached the network. In the AdEx setting, temporal behavior develops through the internal evolution of $V$ and $w$ during network processing. 

This distinction is central to the privacy question considered here. If privacy differences persist when temporal diversity is generated through the neuron dynamics themselves, then the observed effect cannot be attributed solely to how spikes are arranged at the input. Instead, the membrane and adaptation mechanisms that govern how activity evolves within the network may also contribute to the resulting privacy--utility behavior. AdEx therefore provides a controlled way to examine whether richer intrinsic dynamics introduce privacy characteristics that are absent, or less pronounced, under conventional LIF dynamics.
\begin{figure*}[t]
    \centering
    \includegraphics[
        width=\linewidth,
        trim={0 30mm 0 32mm},
        clip
    ]{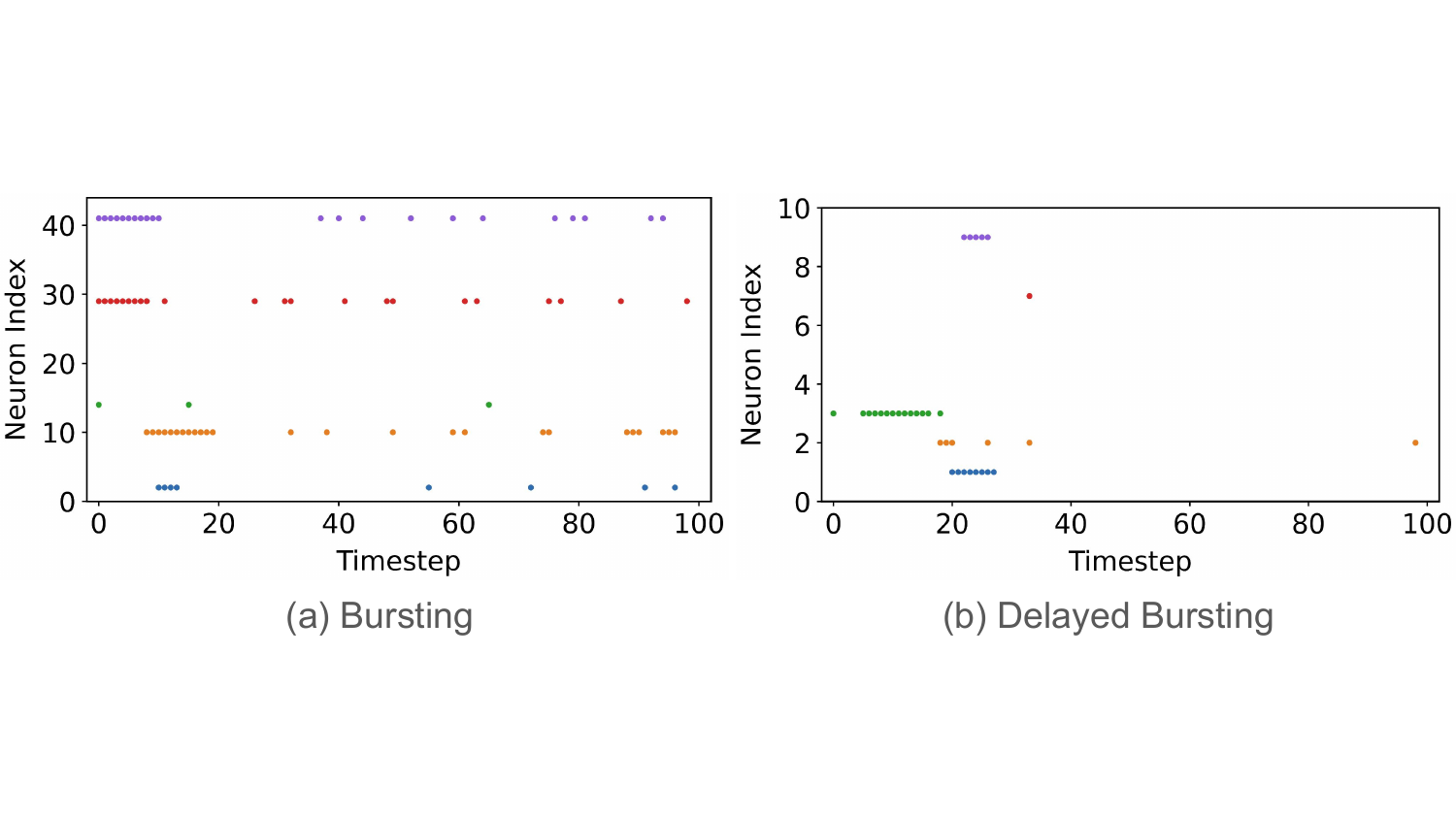}
    \caption{Representative AdEx spike rasters for the Burst and Delayed-Burst configurations. The Burst setting produces temporally clustered spike activity, whereas the Delayed-Burst setting shifts concentrated firing to later time steps and results in sparser activation.}
    \label{fig:adex_raster}
\end{figure*}

\subsubsection{\textbf{Experimental Design for Intrinsic Neuron Dynamics}}

We evaluate three AdEx firing regimes: Rate, Burst, and Delayed Burst. The Rate configuration serves as the reference condition and produces comparatively regular spiking. The Burst and Delayed-Burst configurations are obtained by modifying the threshold, adaptation, and reset mechanisms that determine spike initiation and the temporal evolution of activity following previous spikes \cite{naud2008firing}.

For the Burst configuration, the effective firing threshold is set to $0.45$, with a relatively short adaptation time constant of $\tau_w=30$ and weak subthreshold adaptation $(a=0.005)$. The post-spike reset is set to $V_{\mathrm{reset}}=0.30$, while the spike-triggered adaptation increment is $b=0.08$. This parameterization promotes repeated firing over short temporal intervals. The corresponding raster in Figure~\ref{fig:adex_raster}(a) shows this behavior as dense groups of closely spaced spikes across several neurons, particularly during the early portion of the simulation. Similar clustered events also reappear at later time steps, which indicates that the Burst configuration supports repeated episodes of temporally concentrated activity rather than isolated firing events.
\begin{figure*}[t]
    \centering
    \includegraphics[
        width=\linewidth,
        trim={0 0mm 0 0mm},
        clip
    ]{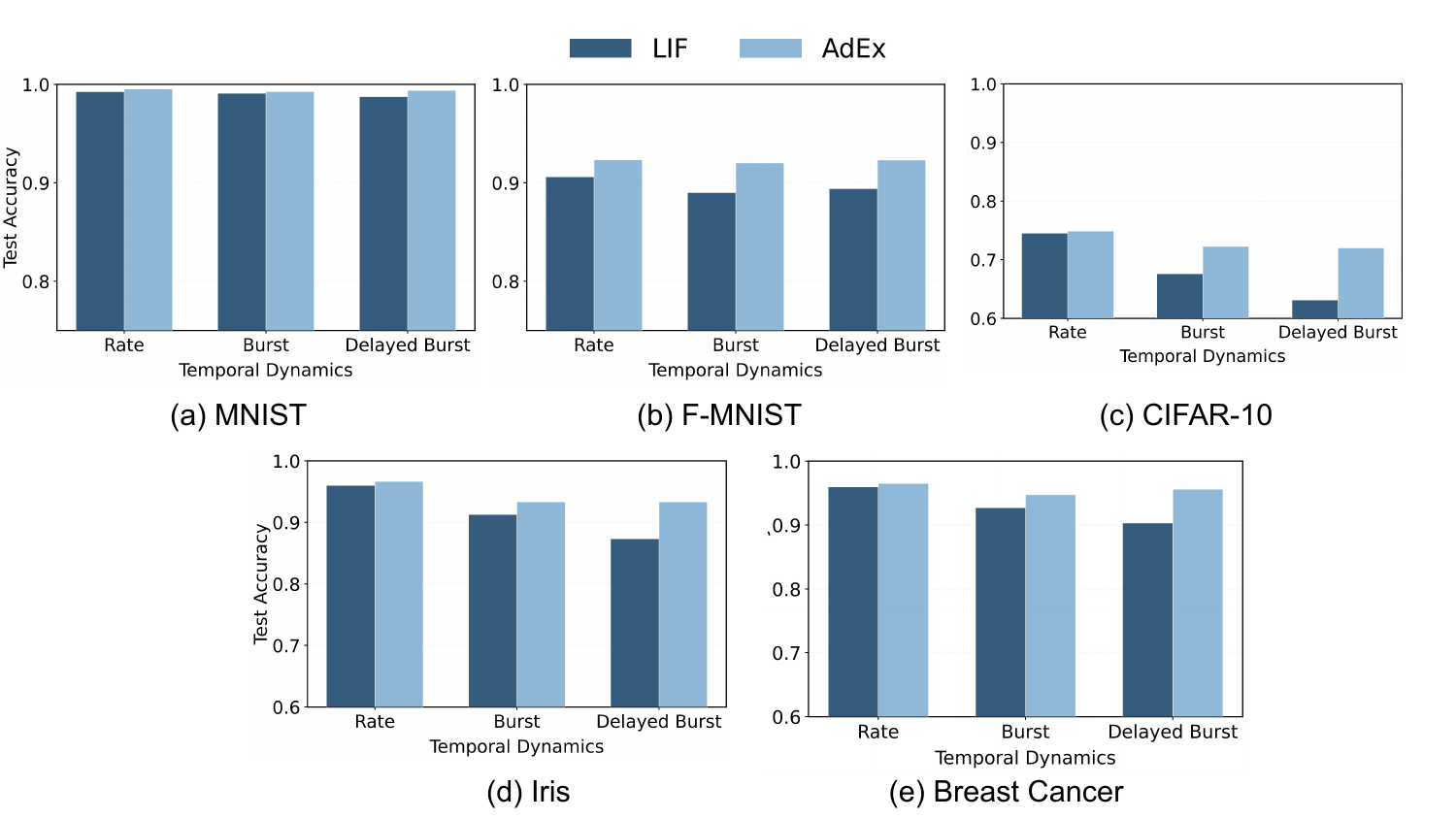}
    \caption{Classification utility of LIF and AdEx neurons under Rate, Burst, and Delayed-Burst temporal dynamics on (a) MNIST, (b) F-MNIST, (c) CIFAR-10, (d) Iris, and (e) Breast Cancer.}
    \label{fig:adex_perf}
\end{figure*}

\begin{figure*}[t]
    \centering
    \includegraphics[
        width=\linewidth,
        trim={0 0mm 0 0mm},
        clip
    ]{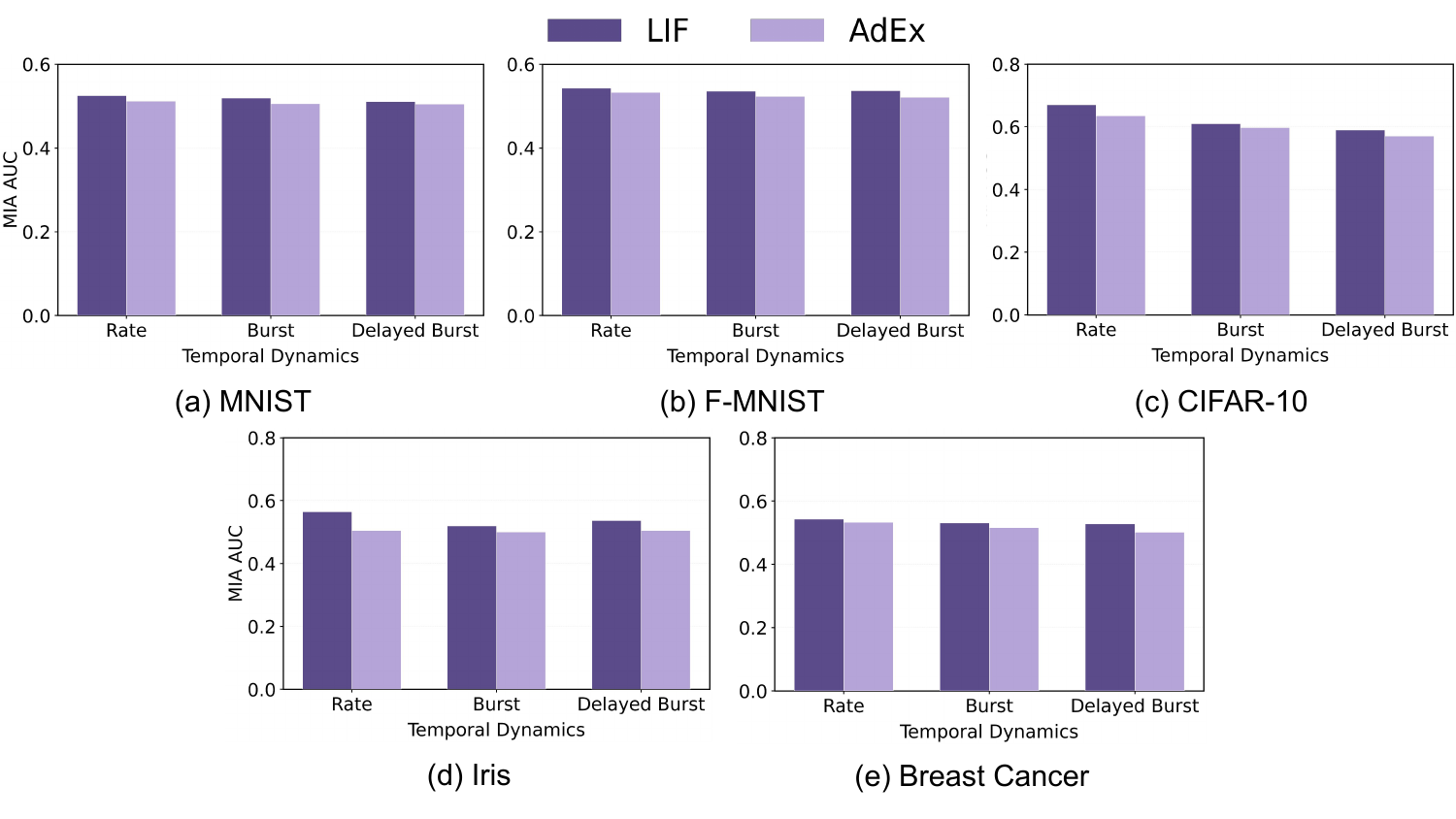}
    \caption{Membership-inference vulnerability (MIA AUC) of LIF and AdEx neurons under Rate, Burst, and Delayed-Burst temporal dynamics on (a) MNIST, (b) F-MNIST, (c) CIFAR-10, (d) Iris, and (e) Breast Cancer. Lower values indicate reduced leakage.}
    \label{fig:adex_mia}
\end{figure*}

\begin{figure*}[t]
    \centering
    \includegraphics[
        width=\linewidth,
        trim={0 18mm 0 20mm},
        clip
    ]{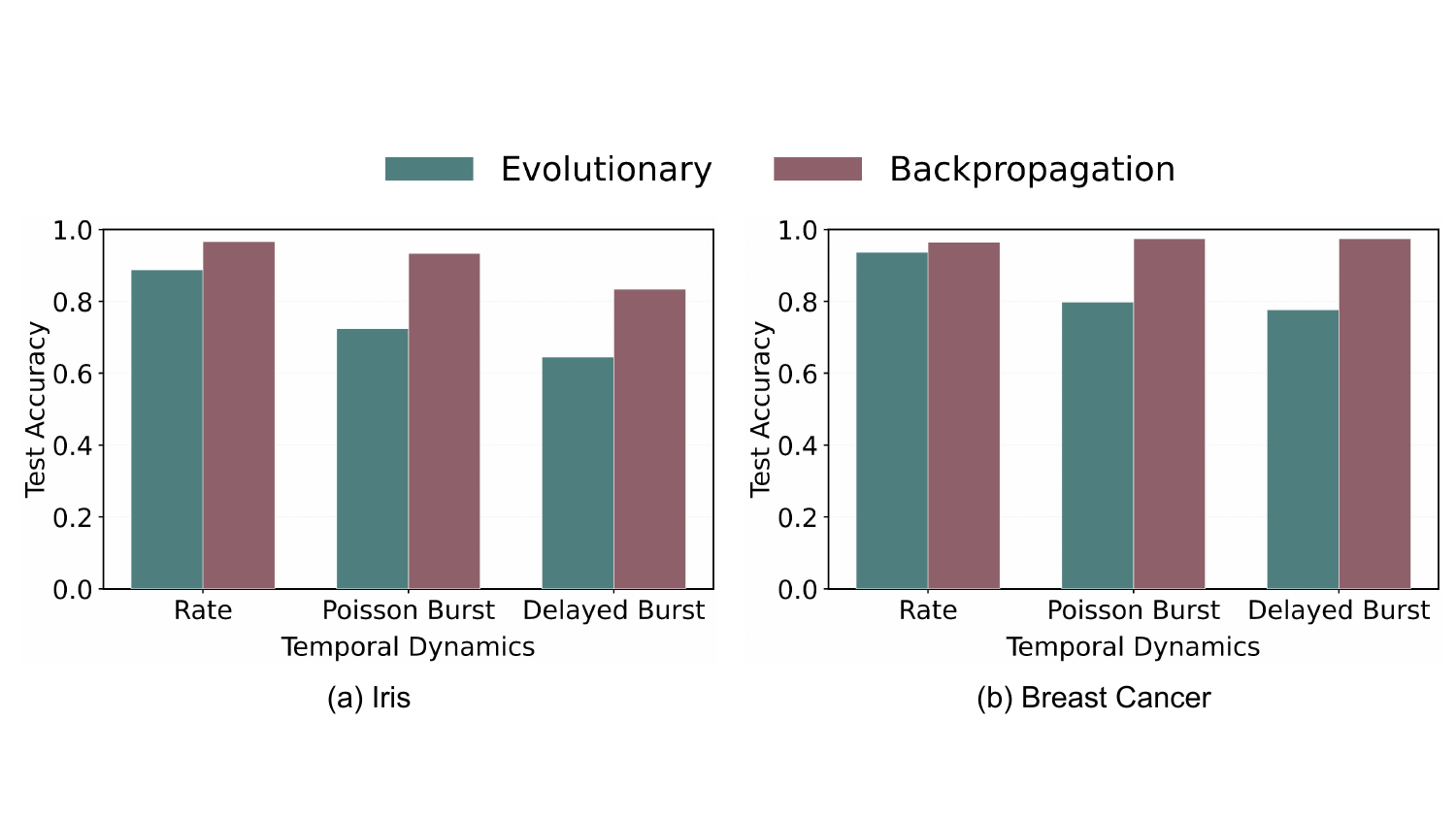}
    \caption{Classification utility of SNNs trained with evolutionary learning and backpropagation under Rate, Poisson-Burst, and Delayed-Burst temporal dynamics on (a) Iris and (b) Breast Cancer.}
    \label{fig:alg_perf}
\end{figure*}
The Delayed-Burst configuration increases the firing threshold to $0.54$ and extends the adaptation time constant to $\tau_w=90$. Subthreshold adaptation is strengthened to $a=0.03$, the spike-triggered adaptation increment is increased to $b=0.10$, and the reset potential is set to $V_{\mathrm{reset}}=0.34$. These changes reduce the tendency toward immediate repeated firing and shift concentrated activity later in the temporal sequence. As shown in Figure~\ref{fig:adex_raster}(b), the resulting activity is substantially sparser, with several neurons remaining inactive during the initial time steps before short clusters emerge later in the sequence. The delayed configuration therefore produces greater firing latency and a more restricted temporal concentration of activity, rather than uniformly delaying every neuron by the same amount.

The two AdEx configurations therefore induce distinct temporal response regimes within the network. This separation permits direct evaluation of whether intrinsic differences in spike-generation dynamics are associated with corresponding changes in utility and membership leakage.

\subsubsection{\textbf{Result and Analysis}}

Figure~\ref{fig:adex_perf} shows that replacing LIF with AdEx preserves or improves classification performance across the evaluated datasets and temporal dynamics. The difference is small on MNIST, where both neuron models already achieve high accuracy, but becomes more pronounced on F-MNIST, CIFAR-10, Iris, and Breast Cancer. The improvement is particularly evident under Burst and Delayed-Burst dynamics, where AdEx recovers a substantial portion of the utility lost with LIF. This pattern suggests that the additional adaptation and membrane dynamics of AdEx allow the network to support more complex temporal activity without incurring the same degree of performance degradation.

The privacy results in Figure~\ref{fig:adex_mia} show a similarly consistent advantage for AdEx. Across the five datasets, AdEx produces lower MIA AUC than LIF under Rate, Burst, and Delayed-Burst dynamics. The magnitude of the reduction varies by dataset and temporal condition, but the direction of the effect remains largely consistent. The improvement is particularly visible on CIFAR-10, Iris, and Breast Cancer, where AdEx lowers membership leakage while simultaneously maintaining or improving classification performance.

The combined utility and privacy behavior indicates that intrinsic neuron dynamics influence the privacy--utility characteristics of SNNs independently of the temporal pattern presented to the network. The privacy differences observed under AdEx are not accompanied by the utility degradation seen when temporal structure is altered through external perturbation alone. Instead, richer internal neuronal dynamics can reshape how temporal information is processed in a way that improves task performance while also reducing membership leakage. This supports the view that privacy in SNNs depends not only on the temporal structure of spike activity, but also on the neuronal mechanisms that generate and transform that activity.

\section{Learning and Training Mechanisms in Privacy}

\subsection{\textbf{Privacy under Gradient-Based and Evolutionary Learning}}
\label{sec:learning}

The learning algorithm determines how individual training samples shape the network parameters and, in turn, how much sample-specific information the trained model retains. Gradient-based training updates parameters using loss derivatives computed directly on the training data. Evolutionary learning instead explores a population of candidate networks and selects among them using a scalar, task-level fitness, without access to gradient information. Because the two paradigms couple the training data to the model in fundamentally different ways, they may yield different membership-leakage profiles even when the temporal representation is held constant. We therefore evaluate Rate, Poisson-Burst, and Delayed-Burst dynamics under both paradigms to determine whether the privacy behavior observed under backpropagation is specific to gradient-based optimization or persists under a fundamentally different learning process.

Evolutionary optimization is implemented using EONS within the TENNLab framework~\cite{plank2018tennlab}. This choice follows our earlier study~\cite{moshruba2024neuromorphic}, in which evolutionary SNNs yielded MIA AUC values close to 0.5 across variations in population size, mutation rate, crossover rate, and processor configuration, and exhibited lower leakage than surrogate-gradient models on Iris and Breast Cancer. As described in Section~\ref{sec:learning-algorithms}, the evolutionary experiments are restricted to these two datasets because of the computational cost of evaluating entire populations. The three temporal dynamics are incorporated into the EONS pipeline as follows: Rate dynamics uses the native TENNLab encoder output, whereas Poisson-Burst and Delayed-Burst spike trains are generated from the firing probabilities derived from this encoder, following the same temporal definitions used in the preceding experiments.

For each temporal configuration, the target and shadow SNNs are evolved independently, with classification accuracy as the fitness objective. Each shadow model is evolved on its corresponding data subset using the same procedure as the target model, after which the standard shadow-model MIA pipeline is applied. The temporal representation and the attack procedure are therefore held fixed across learning paradigms, so that differences in utility and membership leakage can be attributed primarily to the learning algorithm.

\subsubsection{\textbf{Result and Analysis}}

Figure~\ref{fig:alg_perf} shows that Rate dynamics yield the highest or near-highest accuracy under both learning paradigms, but the accuracy cost of introducing burst-based dynamics depends strongly on the learning paradigm. Under evolutionary learning, accuracy decreases consistently from Rate to Poisson-Burst to Delayed-Burst. On Iris, Delayed-Burst results in a 27.3\% relative accuracy degradation from the Rate baseline, compared with 13.8\% under backpropagation. On Breast Cancer, backpropagation retains its accuracy under both burst-based dynamics. As discussed in Section~\ref{sec:homogeneous}, burst-based dynamics make temporal integration more difficult for the downstream LIF network. Backpropagation can compensate for this difficulty because it uses gradient information across the simulation window to adjust the weights to the temporal pattern of the input spikes. Evolutionary learning has no access to gradient information and relies on task-level fitness alone, so it compensates less effectively. The learning paradigm therefore determines how well the network handles burst-based dynamics.

\begin{figure*}[t]
    \centering
    \includegraphics[
        width=\linewidth,
        trim={0 18mm 0 20mm},
        clip
    ]{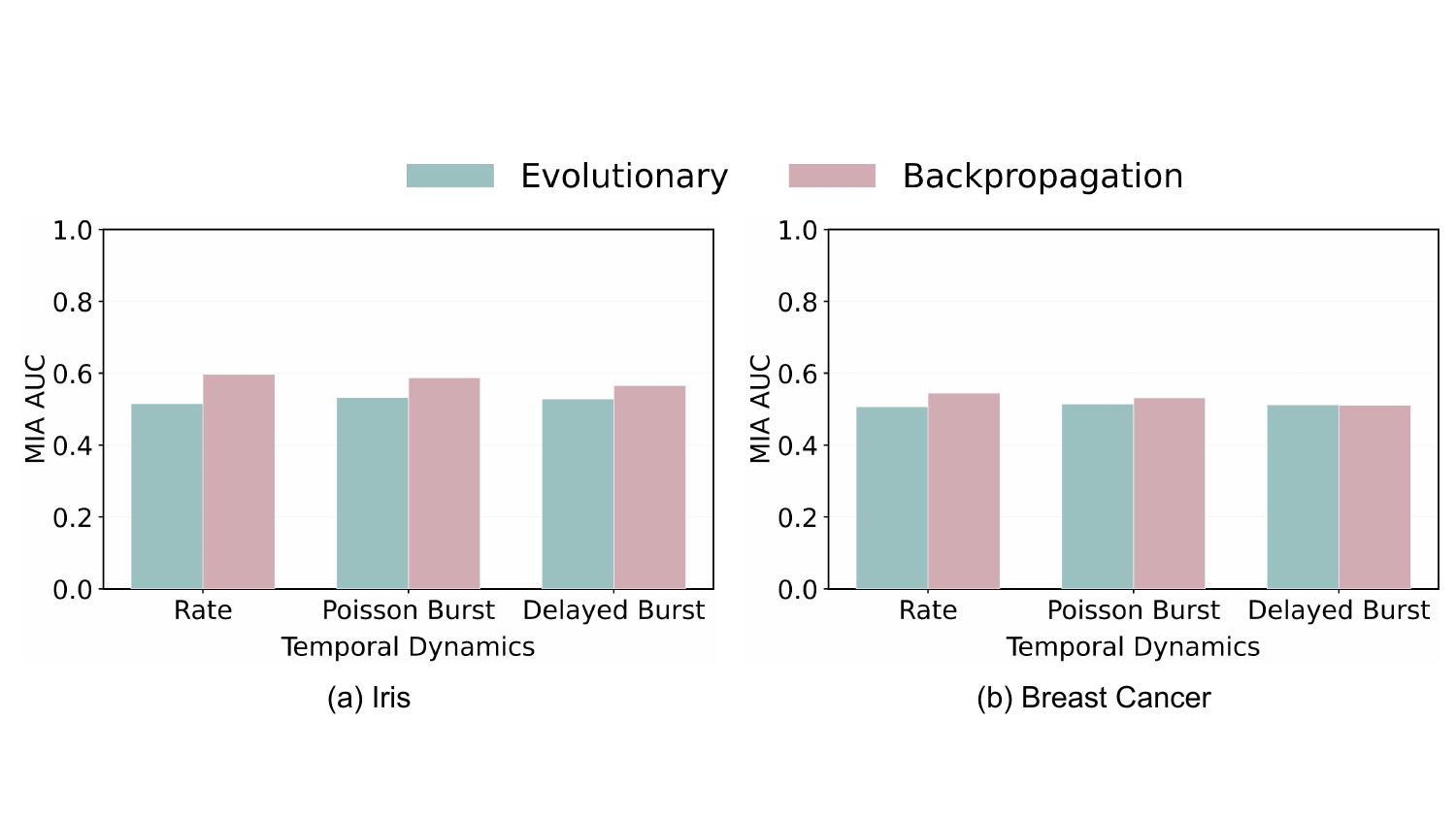}
    \caption{Membership-inference vulnerability of SNNs trained with evolutionary learning and backpropagation under Rate, Poisson-Burst, and Delayed-Burst temporal dynamics on (a) Iris and (b) Breast Cancer.}
    \label{fig:alg_mia}
\end{figure*}

Figure~\ref{fig:alg_mia} shows that under backpropagation, MIA AUC decreases progressively from Rate to Poisson-Burst to Delayed-Burst on both datasets, consistent with Section~\ref{sec:homogeneous}. Under evolutionary learning, MIA AUC remains close to 0.5 across all temporal dynamics. Because the Rate baseline already yields an MIA AUC near 0.5, which corresponds to random guessing, little reduction in leakage can be observed in this setting. The evolutionary results are therefore inconclusive with respect to the effect of burst-based dynamics on membership leakage.

Considering accuracy and leakage, Figures~\ref{fig:alg_perf} and~\ref{fig:alg_mia} indicate that the trade-off introduced by burst-based dynamics is conditioned by the learning paradigm. Burst-based temporal dynamics improve privacy under backpropagation. Under evolutionary learning, the Rate baseline already exhibits near-ideal privacy, so the effect of burst-based dynamics is confined to utility, where they cause a substantial degradation. The privacy benefit of burst-based dynamics is therefore contingent on the baseline leakage established by the learning paradigm.

\subsection{\textbf{Differentially Private Training}}

Differentially Private Stochastic Gradient Descent (DPSGD) provides a formal mechanism for limiting the influence of individual training samples during optimization. DPSGD clips per-sample gradients and adds calibrated noise before the parameter update, which bounds the contribution of any individual record to the learned model \cite{abadi2016deep}. The resulting privacy guarantee is characterized by $(\epsilon,\delta)$, where $\epsilon$ represents the privacy budget. A smaller $\epsilon$ imposes a stronger privacy guarantee and requires greater perturbation during training, which can reduce predictive utility. As $\epsilon$ increases, the privacy constraint is relaxed, less perturbation is required, and classification performance can recover. The variation of utility across privacy budgets therefore provides a direct measure of the privacy--utility tradeoff.

We use DPSGD to examine how this tradeoff interacts with Rate, Poisson-Burst, and Delayed-Burst temporal dynamics. The objective is to determine whether the temporal dynamics retain different utility characteristics when the level of privacy protection is explicitly controlled, and whether burst-based dynamics respond differently to the optimization noise introduced by DPSGD.

DPSGD is implemented using Opacus \cite{yousefpour2021opacus}. We evaluate privacy budgets from $\epsilon=0.5$ to $2.0$, with $\delta=10^{-5}$ and a maximum gradient norm of 1.0. For each target $\epsilon$, Opacus calibrates the required noise level while the SNN architecture and temporal-dynamics configuration are otherwise maintained across experiments. 

\subsubsection{\textbf{Result and Analysis}}

\begin{figure*}[t]
    \centering
    \includegraphics[
        width=\linewidth,
        trim={0 14mm 0 0mm},
        clip
    ]{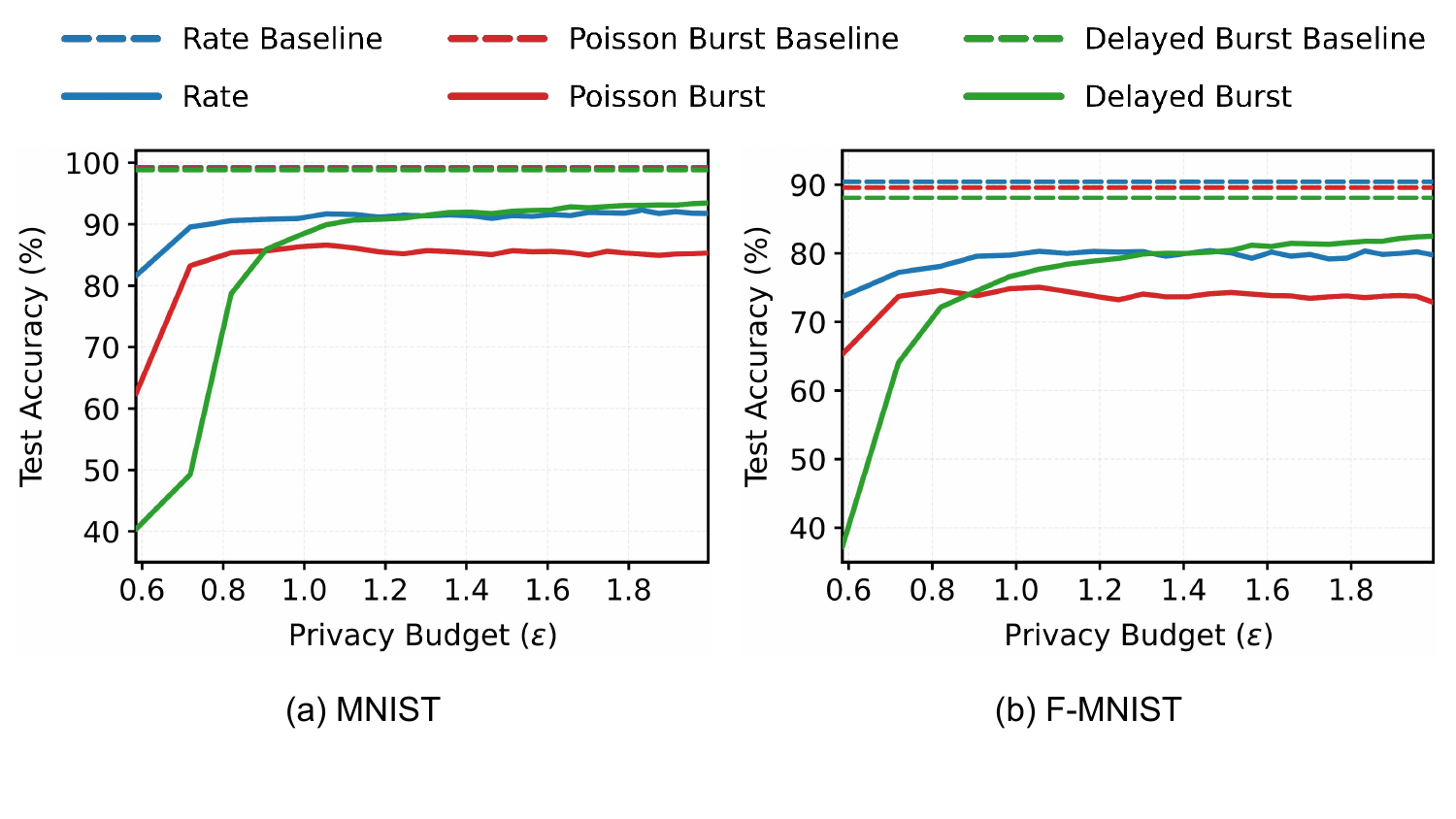}
    \caption{Classification performance of Rate, Poisson-Burst, and Delayed-Burst temporal dynamics under increasing DPSGD privacy budgets on (a) MNIST and (b) F-MNIST. Solid lines show test accuracy as $\epsilon$ increases, while dashed lines indicate the corresponding non-private baseline accuracy for each temporal dynamic.}
    \label{fig:dpsgd}
\end{figure*}

Figure~\ref{fig:dpsgd} shows that, at an identical privacy budget, the temporal dynamics retain different amounts of accuracy. At strict budgets, the non-private ordering is preserved, with Delayed-Burst the most affected. As $\epsilon$ increases, Delayed-Burst recovers to the accuracy of Rate on both datasets, whereas Poisson-Burst remains approximately 7 percentage points below Rate, a deficit larger than that observed without privacy constraints. Since $\epsilon$ imposes the same privacy guarantee on all three dynamics, these differences reflect the utility that each dynamic sacrifices to attain equal protection.

The origin of this cost becomes apparent when DPSGD is compared with the perturbations of Section~\ref{sec:stochasticity}. There, noise corrupted the spike representation itself, and burst-based dynamics lost more accuracy than Rate because the task-relevant information they carry resides in the temporal organization of spikes. DPSGD instead perturbs only the parameter updates and leaves the representation intact. Under this perturbation, Delayed-Burst, whose onset delay is bounded, matches Rate once the noise is reduced, which indicates that its earlier sensitivity was tied to corruption of the representation and not to the dynamic itself. Poisson-Burst, in which both burst occurrence and burst size are random, remains below Rate in both settings, which suggests that its intrinsic stochasticity compounds with the injected noise regardless of where the latter enters.

Together with Sections~\ref{sec:stochasticity} and~\ref{sec:learning}, this shows that the effect of noise on a temporal dynamic depends on the stage at which it is introduced, the representation or the learning process, and on the intrinsic stochasticity of the dynamic on which it acts. The privacy--utility behavior of temporal dynamics is therefore an interaction effect and not an intrinsic property of the dynamics, consistent with the conclusion that privacy in SNNs emerges from the interaction among temporal dynamics, stochasticity, and learning.

\section{Discussion}

The temporal dynamics of SNNs determine how information about each input is represented and retained over time, and changes in these dynamics alter membership leakage even when the network architecture remains fixed. This shows that privacy in SNNs is not a fixed architectural property. The central issue is where this variation originates. A shift from Rate to Poisson-Burst or Delayed-Burst modifies several properties simultaneously, including spike timing, burst organization, stochasticity, and the temporal distribution of activity. The resulting privacy differences may arise from any of these factors, or from their interaction with neuronal dynamics and the learning process. The experiments in this study were designed to progressively separate these effects and identify the mechanisms that shape the observed leakage behavior.

The initial comparison of homogeneous and heterogeneous temporal dynamics provides the first evidence that temporal organization itself contributes to the observed privacy variation. Under a common LIF architecture and learning procedure, Rate, Poisson-Burst, and Delayed-Burst produce clearly different privacy-utility profiles. Burst-based dynamics generally reduce membership leakage relative to Rate, but the reduction is accompanied by lower classification accuracy. Introducing temporal heterogeneity further shows that privacy does not improve simply by increasing the amount of burst-based activity. Moderate mixtures can reduce leakage while retaining much of the utility of Rate dynamics, whereas stronger heterogeneity can reduce accuracy without producing a corresponding privacy gain. The privacy effect is therefore linked to how information is distributed across temporal patterns, rather than to the presence of bursting alone. 

One possible origin of this behavior is the additional stochasticity introduced by the burst-based dynamics. If randomness were the dominant source of privacy, then deliberately increasing stochastic variation should produce a consistent reduction in membership leakage. The perturbation experiments do not support that interpretation. Temporal jitter reduces timing precision and often decreases utility, yet its effect on leakage is non-monotonic. Input noise can reduce leakage in some settings, but the three temporal dynamics continue to exhibit different responses as the perturbation increases. Spike-level noise produces somewhat stronger reductions under heavy corruption, but the effect remains dependent on the underlying temporal structure. Across these settings, increasing stochasticity changes privacy, but does not reproduce a common or progressively stronger privacy effect. The results therefore separate stochasticity from the broader temporal mechanism: randomness contributes to leakage behavior, but it does not account for the privacy differences among the temporal dynamics by itself.

The manner in which perturbation enters the system provides an additional distinction. Input noise alters the signal before temporal activity is formed, temporal jitter changes the timing of existing spikes, and spike noise directly modifies the generated spike sequence. DPSGD introduces variability at a different stage by perturbing the optimization process while leaving the temporal representation intact. These interventions do not produce equivalent responses. Delayed-Burst is highly sensitive under strict DPSGD budgets but recovers strongly as the privacy constraint is relaxed, whereas Poisson-Burst retains a larger utility deficit. The same temporal dynamic can therefore respond differently depending on whether variability acts on the representation itself or on the process through which that representation is learned. Privacy-related effects cannot be attributed to noise magnitude alone. They depend on the computational stage at which the perturbation interacts with temporal information.

Temporal structure also does not act independently of the neuron that processes it. The comparison between LIF and AdEx shows that richer intrinsic neuronal dynamics change both utility and membership leakage across the evaluated firing regimes. AdEx consistently reduces leakage relative to LIF while preserving or improving classification performance. This result is important because the source of temporal behavior is no longer confined to the input representation. Under AdEx, membrane evolution and adaptation shape how activity develops within the network itself. The privacy behavior of a firing regime therefore depends not only on the temporal pattern presented at the input, but also on the internal dynamics through which that pattern is transformed. This places intrinsic neuron dynamics alongside temporal organization as a distinct contributor to the privacy-utility behavior of the network. 

The dependence becomes even clearer when the learning mechanism is changed. Under backpropagation, Poisson-Burst and Delayed-Burst generally reduce membership leakage relative to Rate. Under evolutionary learning, the Rate condition already produces MIA performance close to random guessing, leaving little observable privacy improvement for the burst-based dynamics while their utility continues to decline. The privacy advantage associated with a temporal dynamic is therefore conditioned by how the model learns from individual samples. Gradient-based optimization directly propagates sample-dependent error information through the temporal computation, whereas evolutionary optimization selects candidate networks through task-level fitness without the same form of gradient-based sample coupling. These different mechanisms alter the amount and form of training-specific information retained by the resulting model. A temporal dynamic cannot therefore be assigned a fixed privacy characteristic independently of the learning process in which it is used. 

Viewed together, the results support a more complete account of where membership privacy originates in these SNNs. Temporal dynamics determine how task information is distributed across spike events and time. Stochasticity perturbs that representation, but its effect depends on where it is introduced and on the temporal structure already present. Intrinsic neuron dynamics determine how temporal activity is integrated and transformed as it propagates through the network. The learning mechanism determines how information from individual samples shapes the final model. Membership leakage emerges from the interaction among these processes. The lower leakage observed for a particular temporal dynamic is therefore not an intrinsic property of that pattern, but the outcome of how that pattern is generated, processed, and learned within a specific computational configuration.

This interpretation also changes how privacy-aware SNNs should be designed. Selecting a temporal dynamic solely because it produces lower MIA AUC in one experimental setting is unlikely to generalize across neuron models or training procedures. The heterogeneous results show that privacy gains can arise without completely replacing Rate dynamics, while the AdEx experiments show that modifying intrinsic neuronal behavior can improve privacy without imposing the same utility loss observed when stronger temporal structure is introduced only at the input. The learning experiments further show that a temporal modification can provide substantial privacy benefit in one optimization regime and little additional benefit in another. Privacy-aware design should therefore treat temporal dynamics, neuronal dynamics, and learning as coupled design dimensions. The objective is not to maximize bursting, stochasticity, or temporal complexity independently, but to identify configurations that reduce sample-specific distinguishability while preserving task-relevant temporal information.

The present findings are bounded by the scope of the evaluated models and attacks. Privacy is measured through a consistent black-box shadow-model MIA, which enables controlled comparison but does not characterize all forms of information leakage. The analysis also considers a limited set of temporal dynamics and two neuron models, and evolutionary learning is evaluated only on the smaller tabular datasets because of its computational cost. These constraints leave an important next step: determining how membership information develops internally across time and network depth. Tracking the separation of member and non-member representations through successive temporal and neuronal states could reveal where the leakage signal first emerges and how different mechanisms suppress or amplify it. Extending the same analysis to additional neuron models, learning rules, neuromorphic datasets, and attack strategies would further establish the extent to which the interaction identified here governs privacy across SNN systems.
\section{Conclusion}

This work investigated the origins of membership leakage in SNNs by systematically separating the contributions of temporal representation, stochasticity, intrinsic neuron dynamics, and learning. The results show that privacy in SNNs is not a fixed property of spike-based computation or of any single temporal pattern. Biologically inspired temporal dynamics alter membership leakage through the way they distribute task information across spike events and time, but when imposed at the input they reduce leakage at a cost in accuracy. Stochasticity influences leakage without accounting for the differences among temporal dynamics, and its effect depends on the stage at which it enters the computation. Intrinsic neuron dynamics act as a distinct contributor, as adaptive neurons reduce leakage while preserving or improving classification performance. The learning mechanism further determines whether a temporal dynamic yields any observable privacy benefit. Membership leakage therefore emerges from the interaction among how temporal activity is generated, processed, and learned within a given computational configuration. This understanding provides a principled basis for privacy-aware neuromorphic design, in which designers select temporal dynamics, neuronal dynamics, and learning mechanisms jointly rather than in isolation. Future work will examine how membership information develops across time and network depth. It will also extend this analysis to additional neuron models, learning rules, neuromorphic datasets, and attack strategies, with the goal of establishing design principles for spiking systems that remain both efficient and private in sensitive applications.

\section*{Funding}
This work was supported by the National Science Foundation under Award Nos. 2539714 and 2319619, and by the U.S. Department of Energy under Award No. DE-SC0026260.

\bibliographystyle{unsrt}  
\bibliography{references}

\begin{thebibliography}{10}

\bibitem{davies2018loihi}
Mike Davies, Narayan Srinivasa, Tsung-Han Lin, Gautham Chinya, Yongqiang Cao,
  Sri~Harsha Choday, Georgios Dimou, Prasad Joshi, Nabil Imam, Shweta Jain,
  Yuyun Liao, Chit-Kwan Lin, Andrew Lines, Ruokun Liu, Deepak Mathaikutty,
  Steven McCoy, Arnab Paul, Jonathan Tse, Guruguhanathan Venkataramanan,
  Yi-Hsin Weng, Andreas Wild, Yoonseok Yang, and Hong Wang.
\newblock Loihi: A neuromorphic manycore processor with on-chip learning.
\newblock {\em IEEE Micro}, 38(1):82--99, 2018.

\bibitem{yang2022lead}
Helin Yang, Kwok-Yan Lam, Liang Xiao, Zehui Xiong, Hao Hu, Dusit Niyato, and
  H.~Vincent Poor.
\newblock Lead federated neuromorphic learning for wireless edge artificial
  intelligence.
\newblock {\em Nature Communications}, 13:4269, 2022.

\bibitem{zhu2024autonomous}
Rui-Jie Zhu, Ziqing Wang, Leilani Gilpin, and Jason~K. Eshraghian.
\newblock Autonomous driving with spiking neural networks.
\newblock In {\em Advances in Neural Information Processing Systems}, 2024.

\bibitem{halaly2023autonomous}
Raz Halaly and Elishai Ezra~Tsur.
\newblock Autonomous driving controllers with neuromorphic spiking neural
  networks.
\newblock {\em Frontiers in Neurorobotics}, 17:1234962, 2023.

\bibitem{tian2023neurocare}
Fengshi Tian, Jie Yang, Shiqi Zhao, and Mohamad Sawan.
\newblock Neurocare: A generic neuromorphic edge computing framework for
  healthcare applications.
\newblock {\em Frontiers in Neuroscience}, 17:1093865, 2023.

\bibitem{chu2022neuromorphic}
Haoming Chu, Yulong Yan, Leijing Gan, Hao Jia, Liyu Qian, Yuxiang Huan, Lirong
  Zheng, and Zhuo Zou.
\newblock A neuromorphic processing system with spike-driven snn processor for
  wearable ecg classification.
\newblock {\em IEEE Transactions on Biomedical Circuits and Systems},
  16(4):511--523, 2022.

\bibitem{moshruba2024neuromorphic}
Ayana Moshruba, Ihsen Alouani, and Maryam Parsa.
\newblock Are neuromorphic architectures inherently privacy-preserving? an
  exploratory study.
\newblock {\em arXiv preprint arXiv:2411.06613}, 2024.

\bibitem{li2025membership}
Jiaxin Li, Gorka Abad, Stjepan Picek, and Mauro Conti.
\newblock Membership privacy evaluation in deep spiking neural networks.
\newblock In {\em European Symposium on Research in Computer Security}, pages
  207--227. Springer, 2025.

\bibitem{guan2025privacy}
Junyi Guan, Abhijith Sharma, Chong Tian, and Salem Lahlou.
\newblock On the privacy risks of spiking neural networks: A membership
  inference analysis.
\newblock {\em arXiv preprint arXiv:2502.13191}, 2025.

\bibitem{moshruba2025izhikevich}
Ayana Moshruba, Hamed Poursiami, and Maryam Parsa.
\newblock Izhikevich-inspired temporal dynamics for enhancing privacy,
  efficiency, and transferability in spiking neural networks.
\newblock In {\em 2025 International Conference on Neuromorphic Systems
  (ICONS)}, pages 89--97. IEEE, 2025.

\bibitem{tavanaei2019deep}
Amirhossein Tavanaei, Masoud Ghodrati, Saeed~Reza Kheradpisheh, Timoth{\'e}e
  Masquelier, and Anthony~S. Maida.
\newblock Deep learning in spiking neural networks.
\newblock {\em Neural Networks}, 111:47--63, 2019.

\bibitem{auge2021survey}
Daniel Auge, Julian Hille, Etienne Mueller, and Alois Knoll.
\newblock A survey of encoding techniques for signal processing in spiking
  neural networks.
\newblock {\em Neural Processing Letters}, 53:4693--4710, 2021.

\bibitem{eshraghian2023training}
Jason~K. Eshraghian, Max Ward, Emre Neftci, Xinxin Wang, Gregor Lenz, Girish
  Dwivedi, Mohammed Bennamoun, Doo~Seok Jeong, and Wei~D. Lu.
\newblock Training spiking neural networks using lessons from deep learning.
\newblock {\em Proceedings of the IEEE}, 111(9):1016--1054, 2023.

\bibitem{schmutz2020mesoscopic}
Valentin Schmutz, Wulfram Gerstner, and Tilo Schwalger.
\newblock Mesoscopic population equations for spiking neural networks with
  synaptic short-term plasticity.
\newblock {\em The Journal of Mathematical Neuroscience}, 10(1):5, 2020.

\bibitem{burkitt2006review}
Anthony~N. Burkitt.
\newblock A review of the integrate-and-fire neuron model: I. homogeneous
  synaptic input.
\newblock {\em Biological Cybernetics}, 95(1):1--19, 2006.

\bibitem{indiveri2011neuromorphic}
Giacomo Indiveri, Bernab{\'e} Linares-Barranco, Tara~Julia Hamilton, Andr{\'e}
  van Schaik, Ralph Etienne-Cummings, Tobi Delbruck, Shih-Chii Liu, Piotr
  Dudek, Philipp H{\"a}fliger, Sylvie Renaud, Johannes Schemmel, Gert
  Cauwenberghs, John Arthur, Kai Hynna, Fopefolu Folowosele, Sylvain Saighi,
  Teresa Serrano-Gotarredona, Jayawan Wijekoon, Yingxue Wang, and Kwabena
  Boahen.
\newblock Neuromorphic silicon neuron circuits.
\newblock {\em Frontiers in Neuroscience}, 5:73, 2011.

\bibitem{izhikevich2003simple}
Eugene~M. Izhikevich.
\newblock Simple model of spiking neurons.
\newblock {\em IEEE Transactions on Neural Networks}, 14(6):1569--1572, 2003.

\bibitem{brette2005adaptive}
Romain Brette and Wulfram Gerstner.
\newblock Adaptive exponential integrate-and-fire model as an effective
  description of neuronal activity.
\newblock {\em Journal of Neurophysiology}, 94(5):3637--3642, 2005.

\bibitem{neftci2019surrogate}
Emre~O. Neftci, Hesham Mostafa, and Friedemann Zenke.
\newblock Surrogate gradient learning in spiking neural networks: Bringing the
  power of gradient-based optimization to spiking neural networks.
\newblock {\em IEEE Signal Processing Magazine}, 36(6):51--63, 2019.

\bibitem{bittar2022surrogate}
Alexandre Bittar and Philip~N. Garner.
\newblock A surrogate gradient spiking baseline for speech command recognition.
\newblock {\em Frontiers in Neuroscience}, 16:865897, 2022.

\bibitem{naud2008firing}
Richard Naud, Nicolas Marcille, Claudia Clopath, and Wulfram Gerstner.
\newblock Firing patterns in the adaptive exponential integrate-and-fire model.
\newblock {\em Biological Cybernetics}, 99(4--5):335--347, 2008.

\bibitem{shrestha2018slayer}
Sumit~Bam Shrestha and Garrick Orchard.
\newblock Slayer: Spike layer error reassignment in time.
\newblock In {\em Advances in Neural Information Processing Systems},
  volume~31, pages 1412--1421, 2018.

\bibitem{schuman2020eons}
Catherine~D. Schuman, J.~Parker Mitchell, Robert~M. Patton, Thomas~E. Potok,
  and James~S. Plank.
\newblock Evolutionary optimization for neuromorphic systems.
\newblock In {\em Proceedings of the 2020 Annual Neuro-Inspired Computational
  Elements Workshop}, NICE '20, pages 1--9. Association for Computing
  Machinery, 2020.

\bibitem{elbrecht2020evolving}
Daniel Elbrecht, Shruti~R. Kulkarni, Maryam Parsa, J.~Parker Mitchell, and
  Catherine~D. Schuman.
\newblock Evolving ensembles of spiking neural networks for neuromorphic
  systems.
\newblock In {\em 2020 IEEE Symposium Series on Computational Intelligence
  (SSCI)}, pages 1989--1994. IEEE, 2020.

\bibitem{kim2022privatesnn}
Youngeun Kim, Yeshwanth Venkatesha, and Priyadarshini Panda.
\newblock Privatesnn: privacy-preserving spiking neural networks.
\newblock In {\em Proceedings of the AAAI Conference on Artificial
  Intelligence}, volume~36, pages 1192--1200, 2022.

\bibitem{wang2022dpsnn}
Jihang Wang, Dongcheng Zhao, Guobin Shen, Q~Zhang, and Y~Zeng.
\newblock Dpsnn: a differentially private spiking neural network.
\newblock {\em arXiv preprint arXiv:2205.12718}, 1, 2022.

\bibitem{luo2023encryptedsnn}
Xiwen Luo, Qiang Fu, Sheng Qin, and Kaiyang Wang.
\newblock Encrypted-snn: A privacy-preserving method for converting artificial
  neural networks to spiking neural networks.
\newblock In {\em International Conference on Neural Information Processing},
  pages 519--530. Springer, 2023.

\bibitem{nikfam2023homomorphic}
Farzad Nikfam, Raffaele Casaburi, Alberto Marchisio, Maurizio Martina, and
  Muhammad Shafique.
\newblock A homomorphic encryption framework for privacy-preserving spiking
  neural networks.
\newblock {\em Information}, 14(10):537, 2023.

\bibitem{li2023tfhe}
Pengbo Li, Ting Gao, Huifang Huang, Jiani Cheng, Shuhong Gao, Zhigang Zeng, and
  Jinqiao Duan.
\newblock Privacy preserving discretized spiking neural network with tfhe.
\newblock In {\em 2023 International Conference on Neuromorphic Computing
  (ICNC)}, pages 404--413. IEEE, 2023.

\bibitem{li2023fhedicsnn}
Pengbo Li, Huifang Huang, Ting Gao, Jin Guo, and Jinqiao Duan.
\newblock Efficient privacy-preserving convolutional spiking neural networks
  with fhe.
\newblock {\em arXiv preprint arXiv:2309.09025}, 2023.

\bibitem{nikfam2025spyking}
Farzad Nikfam, Alberto Marchisio, Maurizio Martina, and Muhammad Shafique.
\newblock Spyking—privacy-preserving framework for spiking neural networks.
\newblock {\em Frontiers in Neuroscience}, 19:1551143, 2025.

\bibitem{njungle2026privspike}
Nges~Brian Njungle, Erich Jahns, Milan Stojkov, and Michel~A Kinsy.
\newblock Privspike: A privacy-preserving inference framework for deep spiking
  neural networks using homomorphic encryption.
\newblock In {\em International Conference on Applied Cryptography and Network
  Security}, pages 160--188. Springer, 2026.

\bibitem{poursiami2024brainleaks}
Hamed Poursiami, Ihsen Alouani, and Maryam Parsa.
\newblock Brainleaks: On the privacy-preserving properties of neuromorphic
  architectures against model inversion attacks.
\newblock In {\em 2024 International Conference on Machine Learning and
  Applications (ICMLA)}, pages 705--712. IEEE, 2024.

\bibitem{poursiami2025spikes}
Hamed Poursiami, Ayana Moshruba, and Maryam Parsa.
\newblock Do spikes protect privacy? investigating black-box model inversion
  attacks in spiking neural networks.
\newblock In {\em 2025 International Conference on Neuromorphic Systems
  (ICONS)}, pages 34--40. IEEE, 2025.

\bibitem{moshruba2025surrogate}
Ayana Moshruba, Shay Snyder, Hamed Poursiami, and Maryam Parsa.
\newblock On the privacy-preserving properties of spiking neural networks with
  unique surrogate gradients and quantization levels.
\newblock In {\em 2025 International Joint Conference on Neural Networks
  (IJCNN)}, pages 1--8. IEEE, 2025.

\bibitem{luo2025adpsnn}
Xiwen Luo, Qiang Fu, Junxiu Liu, Yuling Luo, Sheng Qin, and Xue Ouyang.
\newblock Encrypted spiking neural networks based on adaptive differential
  privacy mechanism.
\newblock {\em Entropy}, 27(4):333, 2025.

\bibitem{liu2026radpsnn}
Junxiu Liu, Xiwen Luo, Qiang Fu, Yuling Luo, Sheng Qin, and Xue Ouyang.
\newblock Relevance-based adaptive differential private spiking neural
  networks.
\newblock {\em Expert Systems with Applications}, page 129934, 2025.

\bibitem{aitsam2025dpsnn}
Muhammad Aitsam, Samiulhaq Chardiwall, and Alessandro Di~Nuovo.
\newblock Differentially private spiking neural networks: Enhancing privacy.
\newblock In {\em Cybersecurity and Human Capabilities Through Symbiotic
  Artificial Intelligence: Proceedings of the 16th International Conference on
  Global Security, Safety and Sustainability, London}, page 199, 2024.

\bibitem{moshruba2025edgeprivate}
Ayana Moshruba and Maryam Parsa.
\newblock Edge-private spiking graph convolutional learning: Membership
  inference and structural differential privacy.
\newblock In {\em 2025 First International Conference on Intelligent Computing
  and Systems at the Edge (ICEdge)}, volume~1, pages 1--7. IEEE, 2025.

\bibitem{han2023federated}
Bing Han, Qiang Fu, and Xinliang Zhang.
\newblock Towards privacy-preserving federated neuromorphic learning via
  spiking neuron models.
\newblock {\em Electronics}, 12(18):3984, 2023.

\bibitem{aksu2025federated}
Dogukan Aksu, Jesus~Martinez del Rincon, and Ihsen Alouani.
\newblock Privacy in federated learning with spiking neural networks.
\newblock {\em arXiv preprint arXiv:2511.21181}, 2025.

\bibitem{pereira2026sensitivity}
Luiz Pereira, Mirko Perkusich, Dalton Valadares, and Kyller Gorg{\^o}nio.
\newblock On the sensitivity of firing rate-based federated spiking neural
  networks to differential privacy.
\newblock In {\em ICASSP 2026-2026 IEEE International Conference on Acoustics,
  Speech and Signal Processing (ICASSP)}, pages 20357--20361. IEEE, 2026.

\bibitem{liu2026prispark}
Xin Liu, Honglei Yao, Shanjie Xu, Yuhui Jin, Zijie Pan, Jiamin Zheng, and
  Li~Tan.
\newblock Prispark: Privacy-preserving and communication-efficient federated
  spiking neural learning via event-sparse adaptive aggregation.
\newblock {\em Electronics}, 15(15):3311, 2026.

\bibitem{alam2026federated}
Mohammad~Zahangir Alam.
\newblock Privacy-preserving federated spiking neural networks for real-time
  target detection in distributed isac edge systems, January 2026.
\newblock Preprint.

\bibitem{shokri2017membership}
Reza Shokri, Marco Stronati, Congzheng Song, and Vitaly Shmatikov.
\newblock Membership inference attacks against machine learning models.
\newblock In {\em 2017 IEEE Symposium on Security and Privacy (SP)}, pages
  3--18. IEEE, 2017.

\bibitem{yeom2018privacy}
Samuel Yeom, Irene Giacomelli, Matt Fredrikson, and Somesh Jha.
\newblock Privacy risk in machine learning: Analyzing the connection to
  overfitting.
\newblock In {\em 2018 IEEE 31st Computer Security Foundations Symposium
  (CSF)}, pages 268--282. IEEE, 2018.

\bibitem{song2021systematic}
Liwei Song and Prateek Mittal.
\newblock Systematic evaluation of privacy risks of machine learning models.
\newblock In {\em 30th USENIX Security Symposium (USENIX Security 21)}, pages
  2615--2632. USENIX Association, 2021.

\bibitem{carlini2022membership}
Nicholas Carlini, Steve Chien, Milad Nasr, Shuang Song, Andreas Terzis, and
  Florian Tram{\`e}r.
\newblock Membership inference attacks from first principles.
\newblock In {\em 2022 IEEE Symposium on Security and Privacy (SP)}, pages
  1897--1914. IEEE, 2022.

\bibitem{salem2019mlleaks}
Ahmed Salem, Yang Zhang, Mathias Humbert, Pascal Berrang, Mario Fritz, and
  Michael Backes.
\newblock {ML-Leaks}: Model and data independent membership inference attacks
  and defenses on machine learning models.
\newblock In {\em Proceedings of the 26th Annual Network and Distributed System
  Security Symposium (NDSS)}. The Internet Society, 2019.

\bibitem{nasr2019comprehensive}
Milad Nasr, Reza Shokri, and Amir Houmansadr.
\newblock Comprehensive privacy analysis of deep learning: Passive and active
  white-box inference attacks against centralized and federated learning.
\newblock In {\em 2019 IEEE Symposium on Security and Privacy (SP)}, pages
  739--753. IEEE, 2019.

\bibitem{choquettechoo2021label}
Christopher~A. Choquette-Choo, Florian Tramer, Nicholas Carlini, and Nicolas
  Papernot.
\newblock Label-only membership inference attacks.
\newblock In {\em Proceedings of the 38th International Conference on Machine
  Learning}, volume 139 of {\em Proceedings of Machine Learning Research},
  pages 1964--1974. PMLR, 2021.

\bibitem{lecun1998gradient}
Yann LeCun, L{\'e}on Bottou, Yoshua Bengio, and Patrick Haffner.
\newblock Gradient-based learning applied to document recognition.
\newblock {\em Proceedings of the IEEE}, 86(11):2278--2324, 1998.

\bibitem{xiao2017fashion}
Han Xiao, Kashif Rasul, and Roland Vollgraf.
\newblock Fashion-mnist: A novel image dataset for benchmarking machine
  learning algorithms.
\newblock {\em arXiv preprint arXiv:1708.07747}, 2017.

\bibitem{krizhevsky2009learning}
Alex Krizhevsky.
\newblock Learning multiple layers of features from tiny images.
\newblock Technical report, University of Toronto, 2009.

\bibitem{fisher1936use}
Ronald~A. Fisher.
\newblock The use of multiple measurements in taxonomic problems.
\newblock {\em Annals of Eugenics}, 7(2):179--188, 1936.

\bibitem{street1993nuclear}
W.~Nick Street, William~H. Wolberg, and Olvi~L. Mangasarian.
\newblock Nuclear feature extraction for breast tumor diagnosis.
\newblock In {\em Biomedical Image Processing and Biomedical Visualization},
  volume 1905, pages 861--870. SPIE, 1993.

\bibitem{connors1990intrinsic}
Barry~W. Connors and Michael~J. Gutnick.
\newblock Intrinsic firing patterns of diverse neocortical neurons.
\newblock {\em Trends in Neurosciences}, 13(3):99--104, 1990.

\bibitem{gouwens2019classification}
Nathan~W. Gouwens, Staci~A. Sorensen, Jim Berg, Changkyu Lee, Tim Jarsky,
  Jonathan Ting, Susan~M. Sunkin, David Feng, Costas~A. Anastassiou, Eliza
  Barkan, et~al.
\newblock Classification of electrophysiological and morphological neuron types
  in the mouse visual cortex.
\newblock {\em Nature Neuroscience}, 22(7):1182--1195, 2019.

\bibitem{pereznieves2021neural}
Nicolas Perez-Nieves, Vincent C.~H. Leung, Pier~Luigi Dragotti, and Dan F.~M.
  Goodman.
\newblock Neural heterogeneity promotes robust learning.
\newblock {\em Nature Communications}, 12:5791, 2021.

\bibitem{abbott1999lapicque}
L.~F. Abbott.
\newblock Lapicque's introduction of the integrate-and-fire model neuron
  (1907).
\newblock {\em Brain Research Bulletin}, 50(5--6):303--304, 1999.

\bibitem{wang2022linear}
Ziming Wang et~al.
\newblock Linear leaky-integrate-and-fire neuron model based spiking neural
  networks and its mapping relationship to deep neural networks.
\newblock {\em Frontiers in Neuroscience}, 16:857513, 2022.

\bibitem{plank2018tennlab}
James~S. Plank, Catherine~D. Schuman, Grant Bruer, Mark~E. Dean, and Garrett~S.
  Rose.
\newblock The tennlab exploratory neuromorphic computing framework.
\newblock {\em IEEE Letters of the Computer Society}, 1(2):17--20, 2018.

\bibitem{abadi2016deep}
Martin Abadi, Andy Chu, Ian Goodfellow, H.~Brendan McMahan, Ilya Mironov, Kunal
  Talwar, and Li~Zhang.
\newblock Deep learning with differential privacy.
\newblock In {\em Proceedings of the 2016 ACM SIGSAC Conference on Computer and
  Communications Security}, pages 308--318, 2016.

\bibitem{yousefpour2021opacus}
Ashkan Yousefpour, Igor Shilov, Alexandre Sablayrolles, Davide Testuggine,
  Karthik Prasad, Mani Malek, John Nguyen, Sayan Ghosh, Akash Bharadwaj,
  Jessica Zhao, Graham Cormode, and Ilya Mironov.
\newblock Opacus: User-friendly differential privacy library in pytorch.
\newblock {\em arXiv preprint arXiv:2109.12298}, 2021.

\end{thebibliography}

\end{document}